\documentclass[journal]{IEEEtran}
\usepackage{amsthm}
\usepackage{cite}
\usepackage{amsmath}
\usepackage{nccmath}
\usepackage{amssymb}
\usepackage{amsfonts}
\usepackage{graphicx}
\usepackage{fancyhdr}
\usepackage{svg}
\usepackage{bbm}
\usepackage{textcomp}
\usepackage{xcolor}
\usepackage{algorithmicx,algpseudocode}
\usepackage[linesnumbered,ruled]{algorithm2e}
\usepackage{booktabs}
\usepackage{bbold}
\usepackage{kotex}
\usepackage{mwe}
\usepackage{multicol}
\usepackage{multirow}
\usepackage{subfigure}

\theoremstyle{definition}

\usepackage[normalem]{ulem}

\newcommand{\tblue}[1]{{\color{black}{#1}}}

\newcommand{\BfPara}[1]{{\noindent\bf#1.}\xspace}

\begin{document}
\title{Cooperative Multi-Agent Deep Reinforcement Learning for Reliable and Energy-Efficient Mobile Access via Multi-UAV Control}
\author{
    Chanyoung Park, Soohyun Park, 
    Soyi Jung,~\IEEEmembership{Member, IEEE},
    Carlos Cordeiro,~\IEEEmembership{Fellow, IEEE}, 
    and \\
    Joongheon Kim,~\IEEEmembership{Senior Member, IEEE}
    \thanks{This research was funded by National Research Foundation of Korea (2022R1A2C2004869). \textit{(Corresponding authors: Soohyun Park, Soyi Jung)}}
    \thanks{Chanyoung Park, Soohyun Park, and Joongheon Kim are with the School of Electrical Engineering, Korea University, Seoul 02841, Republic of Korea (e-mails: \{cosdeneb,soohyun828,joongheon\}@korea.ac.kr).}
    \thanks{Soyi Jung is with the Department of Electrical of Computer Engineering, Ajou University, Suwon  16499, Republic of Korea (e-mail: sjung@ajou.ac.kr).}
    \thanks{Carlos Cordeiro is with Intel Corporation, Hillsboro, OR 97124 USA (e-mail: carlos.cordeiro@intel.com).}
}
\maketitle

\begin{abstract}
This paper addresses a novel multi-agent deep reinforcement learning (MADRL)-based positioning algorithm for multiple unmanned aerial vehicles (UAVs) collaboration (\textit{i.e.,} UAVs work as mobile base stations). The primary objective of the proposed algorithm is to establish dependable mobile access networks for cellular vehicle-to-everything (C-V2X) communication, thereby facilitating the realization of high-quality intelligent transportation systems (ITS).
The reliable mobile access services can be achieved in following two ways, \textit{i.e.,} \textit{i)} energy-efficient UAV operation and \textit{ii)} reliable wireless communication services.
For energy-efficient UAV operation, the reward of our proposed MADRL algorithm contains the features for UAV energy consumption models in order to realize efficient operations. 
Furthermore, for reliable wireless communication services, the quality of service (QoS) requirements of individual users are considered as a part of rewards and 60\,GHz mmWave radio is used for mobile access. This paper considers the 60\,GHz mmWave access for utilizing the benefits of \textit{i)} ultra-wide-bandwidth for multi-Gbps high-speed communications and \textit{ii)} high-directional communications for spatial reuse that is obviously good for densely deployed users. Lastly, the comprehensive and data-intensive performance evaluation of the proposed MADRL-based algorithm for multi-UAV positioning is conducted in this paper. The results of these evaluations demonstrate that the proposed algorithm outperforms other existing algorithms.
\end{abstract}

\IEEEpeerreviewmaketitle

\section{Introduction}\label{sec:1}
\begin{figure*}[!t]
    \centering
    \includegraphics[width=1\linewidth]{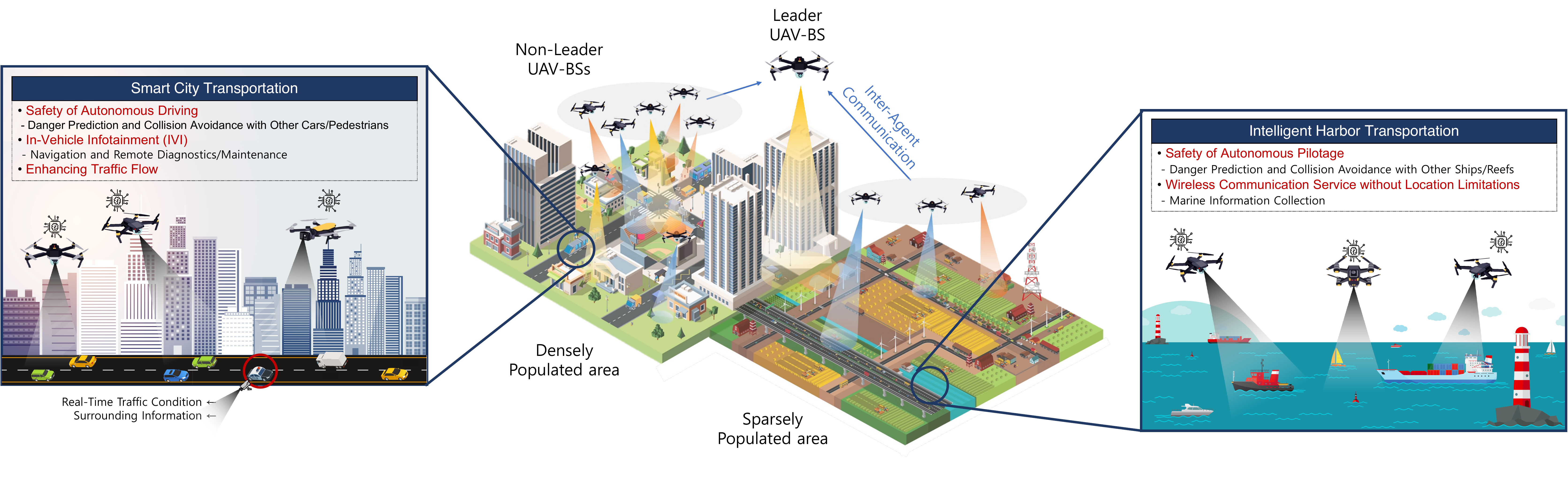}
    \caption{Reference aerial mobile access network model for intelligent transportation systems.}
    \label{fig:system model}
\end{figure*}

\IEEEPARstart{U}{nmanned} aerial vehicles (UAV) have attracted as a part of ad-hoc and flexible infrastructure for mobile and wireless communications~\cite{9044378,tvt201905shin,9428629}. In modern wireless network applications in beyond 5G or 6G networks~\cite{9467353,8869705,tvt202106jung,9358097}, UAVs are widely considered as airborne wireless access points or relay nodes to improve terrestrial communications for instantaneous connectivity services~\cite{9358097}. The UAV-aided mobile network can establish wireless connections without fixed infrastructure, and thus, the network finally realizes broader wireless coverage, achieves higher transmission rates, and provides services in real-time through content caching~\cite{jsac201806choi,cheng2018caching,8471001,twc201912choi,jsac001}.
The deployment of UAV-aided mobile devices can also significantly enhance the functionality of vehicle-to-everything (V2X) communication systems in any transportation setting, including smart cities and harbors~\cite{chen2020vision, 9700761, jrtip21jung} as represented in Fig.~\ref{fig:system model}. By facilitating communication between various traffic-related entities, these networks can provide critical support to autonomous vehicles and intelligent transportation systems (ITS)~\cite{9592698}.
Through enabling vehicular communication via mobile cellular access, autonomous vehicles can obtain real-time information about their surroundings, including the location, speed, and direction of other vehicles. This data exchange can aid in danger prediction and collision avoidance, thereby contributing to the prevention of accidents.
Moreover, real-time updates on traffic conditions, signal timings, and road work information can enhance traffic flow by reducing congestion and improving overall transportation efficiency~\cite{8955944}.
In addition, the benefits of these networks extend beyond road traffic management.
The ability for vehicles to receive and transmit information and services in real-time can be leveraged for various connected services. This includes in-vehicle infotainment (IVI) systems, navigation, and remote diagnostics and maintenance, thereby enhancing the overall user experience and vehicle functionality~\cite{9110903}.
Indeed, the capabilities of existing 5G standards include the provision of services such as ultra-low latency reliable V2X communications. These components offer a fascinating foundation for the establishment of cellular-based operations involving UAVs in the realm of ITS~\cite{americas2018new, ullah20195g}.

Even though the use of UAV-aided mobile networking is beneficial in many applications, the system also has several limitations, \textit{e.g.,} the energy limitation in UAV platforms~\cite{tvt201905shin} and service quality degradation due to unexpected and unmanageable UAV mobility control due to wind gauges and atmospheric conditions. Therefore, it is essential to design and implement an algorithm that is for providing \textit{energy-efficient} and \textit{reliable} mobile access in multi-UAV wireless networks.

For the purpose, machine learning (ML) and deep learning (DL) approaches have become prevalent solutions in the literature~\cite{pieee202105park}. Especially, deep reinforcement learning (DRL) has shown significant performance in sequential stochastic decision-making problems such as resource management for dynamic networking systems such as UAV-aided mobile networks~\cite{lahmeri2021artificial}. Therefore, it is obvious that a novel energy-efficient and reliable algorithm can be designed and implemented using DRL-based methodologies for UAV-aided mobile access services.

In order to design a cooperative and coordinated DRL algorithm for multiple UAV control, multi-agent DRL (MADRL) based algorithm design is essentially required for inter-UAV information-sharing for cooperation and coordination. Among various MADRL algorithms, this paper proposes a new model inspired by \textit{communication neural network (CommNet)} which is fundamentally based on centralized training and distributed execution (CTDE). Therefore, this paper proposes a novel CommNet-based MADRL for determining the optimal positions of UAV base stations (\textit{i.e.,} multi-UAV-BSs positioning) in terms of \textit{i)} cellular network service reliability and \textit{ii)} UAV energy-efficiency. Therefore, the two criteria are formulated as rewards for MADRL formulation. Furthermore, the proposed algorithm considers individual user equipments (UEs)' requirements and characteristics (\textit{e.g.,} mobility).

Moreover, for the cellular network service reliability, 60\,GHz mmWave wireless networks are considered in our proposed reference aerial mobile access network model because they can provide multi-Gbps data rates for \textit{i)} high-speed communications for a better quality of services (QoS) and \textit{ii)} spatial reuse realization for improving per-area data rates due to its high-directionality in wireless transmission~\cite{5733382}. Note that it is obvious that the spatial reuse realization is much more beneficial in urban areas because the UEs are densely deployed.  
Therefore, mmWave wireless communication is widely used in UAV networks in the literature~\cite{mu2,mu3}.

\BfPara{Contributions}
The main contributions of the research in this paper can be summarized as follows.
\begin{itemize}
    \item This paper proposes a new MADRL/CommNet-based cooperative and coordinated multi-UAV-BS positioning algorithm for energy-efficient operations and reliable mobile access services in a distributed manner, under the uncertainty of the environment and the requirements/characteristics of UEs. 
    \item Under the concept of CTDE, our proposed MADRL/CommNet architecture will be trained in a centralized single neural network and the optimization criteria will be based on the reward function of the MADRL formulation. Here, our reward function is formulated based on network service reliability and UAV energy-efficiency. This joint consideration is not previously existing. Furthermore, the reliability can be further improved by utilizing mmWave wireless communications according to ultra-wide bandwidth and high-directionality. This paper is the first MADRL-based UAV research paper which conducts mathematical and fundamental mmWave channel capacity calculation and its practical standard data rate computation with \tblue{modulation and coding scheme (MCS)-based} IEEE 802.11ad standard specification.
    \item Lastly, this paper conducts data-intensive various performance evaluations for the proposed CommNet-based UAV-BS positioning algorithm; and successfully proves that the proposed algorithm outperforms the existing algorithms in various aspects under different environments and experimental settings.
\end{itemize}

\tblue{\subsection{Organization}
The remainder of this paper is structured as follows. Sec.~\ref{sec:related} reviews related work in mobile base station access networks and MADRL algorithms. Sec.~\ref{sec:model} explains our considering network and mmWave communication models. Sec.~\ref{sec:method} presents the proposed autonomous multiple UAV-BS positioning algorithm using CommNet-based MADRL. The novelty of the proposed algorithm is intensively evaluated with various settings in Sec.~\ref{sec:ex}. Lastly, Sec.~\ref{sec:con} concludes this paper.}

\section{Related Work}\label{sec:related}

\subsection{Mobile Access Networks using Multi-UAV Cooperation}\label{sec:uav-bs}
For the reliable aerial cellular network services (via coverage extension and QoS provision) using multiple UAV platforms under the consideration of energy-efficiency operations, it is crucial to deploy the multiple UAV as base stations (\textit{i.e.,} UAV-BSs) into the hotspot area under optimal multi-UAV coordination and cooperation~\cite{9467353}. 
\tblue{This research trends can be classified into two categories, \textit{i.e.,} \textit{i)} ML-based approach (refer to Sec.~\ref{sec:rw1}) and \textit{ii)} control and optimization-based approach (refer to Sec.~\ref{sec:rw2}).}

\subsubsection{ML-based Approach}\label{sec:rw1}
In~\cite{zhang2020predictive, 8937893}, predictive multi-UAV deployment methods are proposed \tblue{using clustering-based unsupervised learning} and the UAVs are considered as temporary BSs to complement ground cellular systems to deal with the cases where managing downlink traffics and providing network coverage extension in emergency and public safety situations.
For providing wireless connectivity for ground UEs, the proposed algorithm in~\cite{liu2018energy} uses an actor-critic based \tblue{DRL} method to solve a multi-objective control problem where the UAVs tend to minimize their energy consumption and extend their coverage.
In addition, the proposed algorithm in ~\cite{9560080} characterizes the fairness at UE-level based on proportional fairness scheduling and formulates a weighted-throughput maximization problem via UAV-BSs' trajectory optimization \tblue{using DRL methodologies}.
Furthermore, a centralized orchestration manager is designed and implemented in order to coordinate a cooperative energy sharing mechanism with MADRL under the consideration of fairness, energy efficiency, and cost-effectiveness, in~\cite{tvt202106jung}. Similarly, an MADRL-based autonomous UAV management method for reliable surveillance in smart city applications was designed and evaluated, where the core idea is autonomously replenishing the UAV's deficient network requirements with the cooperation and coordination among multiple UAV agents~\cite{yun2022cooperative}.

\subsubsection{Control and Optimization-based Approach}\label{sec:rw2}
\tblue{
In~\cite{sargolzaei2020control}, cooperative control techniques for UAVs are reviewed and categorized as consensus control theory, flocking based strategies and formation control. Consensus control involves a collective decision-making process among a group of agents through a sensing or
communication network to reach a common decision. In~\cite{xia2016formation}, an optimal design for consensus control of agents with double-integrator dynamics with collision avoidance considerations is introduced. Similarly, a consensus protocol for a class of high-order multi-agent systems is proposed in~\cite{rezaee2015average}.
In addition, a leader-follower approach for flocking control of multiple UAVs with the goal of conducting a sensing task is introduced in~\cite{quintero2013flocking}. In their strategy, each of the followers is controlled using stochastic optimal control where the cost function takes into account the follower's heading and distance with respect to the leader.
On the other approach, pure pursuit (PP) guidance algorithm is a widely used leader-follower guidance strategy in the formation control which was originally implemented on ground-attack missile systems that aim to hit the target~\cite{lin1991modern}.
}

\subsection{CTDE for MADRL}\label{sec:ctde}
MADRL algorithms and applications can be complex because all given agents potentially interact among each other and learn concurrently. Those interactions constantly reshape the environment and lead to non-stationarity, affecting the trained policy of other agents. In addition, the agents only get a partial observation, \textit{i.e.,} they does not know complete environment information while interacting. It is, therefore, necessary to establish communications among agents during learning when designing MADRL algorithms~\cite{nguyen2020deep}. 
Based on this idea, many MADRL algorithms widely has adopted the concept of CTDE, where it can train a group of agents simultaneously by applying a centralized method via establishing communication channels~\cite{nguyen2020deep}. For more details, CTDE-based agents are trained offline using centralized neural network; and then the CTDE-based agents are executed online after sharing the trained neural network in a decentralized manner~\cite{lowe2017multi}. 
The MADRL can be classified based on communication methods in~\cite{chu2020multi}. The first group is non-communicative and focuses on stabilizing training with advanced value estimation methods. The multi-agent deep deterministic policy gradient (MADDPG) method based on actor-critic (AC) policy gradient algorithms proposed in~\cite{lowe2017multi} features the CTDE paradigm in which the critic uses extra information to ease the training process. At the same time, actors take action based on their local observations. Another multi-agent AC method, \textit{i.e.,} counterfactual multi-agent (COMA)~\cite{foerster2018counterfactual}, is also based on CTDE computation. However, unlike MADDPG, COMA can handle the multi-agent credit assignment problem~\cite{harati2007knowledge} where agents struggle to work their contribution to the team’s success from global rewards generated by the joint actions in cooperative settings. 
Another group proposes learnable communication protocols. Each agent will be connected in this category, and messages are broadcasted, which can be viewed as a fully connected structure, including differentiable inter-agent learning (DIAL)~\cite{foerster2016learning}, bilateral complementary network (BiCNet)~\cite{hou2021bicnet}, and CommNet. As presented in~\cite{sukhbaatar2016learning}, CommNet-based agents learn continuous communications alongside their policy for fully cooperative tasks. CommNet learns a shared centralized neural network for agents to process local observations. Each agent’s decisions depend on observations and the mean vector of messages from other agents.
Notice that recent research contributions have further focused on the open problems of how to communicate more efficiently regarding when, what, and with whom to communicate, based on a predefined or learned communication structure~\cite{zhu2022survey}.

\section{Model}\label{sec:model}
\subsection{ITS-based Mobile Access for Multifaceted Environments}
UAVs not only possess fundamental functionalities such as wireless communications and remote surveillance but also offer advanced features including high mobility, agility, broad accessibility, and rapid deployment~\cite{peer2021usermobility,zeng2019accessing}. These attributes make UAVs a preferred option for diverse purposes within the framework of ITS~\cite{9298484, 9506932, 9190025}.
By leveraging the characteristics of UAV-BSs, aerial wireless networks can be swiftly established, providing essential support to terrestrial cellular networks across a range of scenarios~\cite{mu2}, including those based on ITS~\cite{9312485, 9195789}.
The demand for wireless networks is explosive in densely populated urban areas. However, the cost of establishing terrestrial infrastructures
is high. At this time, mmWave wireless network service within 5G provided by mobile base stations is considered a key technology to meet the demand for wireless communication services in dense urban environments and improve network capacity~\cite{6955989}. For example, UAV-BS can quickly establish a wireless connection with UEs in extreme environments where it is not easy to install terrestrial base stations for technical and economic reasons~\cite{zhang2021joint}. In addition, UAV-BS can be used in urgent situations such as emergency communications in case of disasters and military communications in extreme areas~\cite{bor2016new, alzenad20173, zeng2016wireless}. However, since UAV-BS has a limited communication range, it is challenging and inefficient to always cover all areas with UAV in a dynamic and uncertain environment~\cite{9560080}. Since UAVs are operated by consuming batteries~\cite{gupta2015survey}, battery life must be considered to prevent the UAV from being completely discharged. Designing the optimal UAV-BS trajectories can be alternative solutions to cope with the aforementioned issue~\cite{lu2017energy}.
This paper, we propose a new MADRL algorithm that allows distributed UAV-BSs to cooperatively provide energy-efficient and reliable mmWave wireless communication services by designing the optimal 3D trajectory of UAV-BSs while considering UE mobility in an uncertain and dynamic environment.

\subsection{60\,GHz Millimeter-Wave Model}\label{sec:3-b}
This paper considers 60\,GHz mmWave radio propagation characteristics for mobile UAV-BS access in order to exploit spatial reuse benefits~\cite{6955961}.

\subsubsection{Interference Analysis}
It is important to consider the impact of interference when analyzing the performance of communication systems. To conduct the interference analysis on 60\,GHz mmWave wireless systems, specific path loss models and antenna radiation patterns are considered.

This paper uses the 60\,GHz path loss model defined in the development of the IEEE 802.11ad standard~\cite{maltsev2009path}, \textit{i.e.,} 
\begin{equation}
L(d)=A_{\mathrm{dB}}+20\log_{10}(f_{\mathrm{GHz}})+10\cdot n\cdot \log_{10}(d),
\label{eq:Loss}
\end{equation}
where $A_{dB}=32.5\,\mathrm{dB}$ is an antenna and beamforming specific coefficient. In addition, $f_{\mathrm{GHz}}$, $n$, and $d$ are the carrier frequency in a GHz scale (\textit{i.e.,} $f_{\mathrm{GHz}}=60$), the path loss coefficient ($n=2$), and the distance between a transmitter (Tx) and a receiver (Rx) in a meter scale, respectively. 

The Gaussian reference 60\,GHz antenna radiation pattern is modeled in the International Telecommunication Union (ITU) document~\cite{sector2019reference} as follows,
\begin{equation}\label{eq:radiation}
        G_{\mathrm{dBi}}(\varphi, \theta)=
        \begin{cases}
            G_{\mathrm{dBi}}^{\mathrm{max}}-12\Delta^{2}, & 0\leq \Delta < 1\\
            G_{\mathrm{dBi}}^{\mathrm{max}}-12-15\ln{\Delta}, & 1\leq \Delta
        \end{cases}
\end{equation}
where $G_{\mathrm{dBi}}^{\mathrm{max}}$ is the maximum antenna gain, $\varphi$ and $\theta$ are an azimuth angle and an elevation angle $(-180^{\circ}\leq \varphi \leq 180^{\circ}, -90^{\circ}\leq \theta \leq 90^{\circ})$, respectively. Here, the $\Delta$ can be calculated as $\Delta \triangleq |\Omega^{\dag} \cdot \Omega^{\star}|$, and where $\Omega^{\dag} \triangleq \cos^{-1}{(\cos{\varphi}\cos{\theta})}$ and $\Omega^{\star} \triangleq \sqrt{\left[ \frac{\cos{\left\{\tan^{-1}{\left( \frac{\tan{\theta}}{\sin{\varphi}} \right)}\right\}}}{\varphi_3}\right]^2 +\left[\frac{\sin{\left\{\tan^{-1}{\left( \frac{\tan{\theta}}{\sin{\varphi}} \right)}\right\}}}{\theta_3}\right]^2}$
where $\varphi_3$ and $\theta_3$ are 3\,dB half-power beamwidth in azimuth and elevation planes. This paper assumes $\varphi_3 = \theta_3 = 10^{\circ}$~\cite{kim2017feasibility}.

The interference caused by transmissions on wireless link $i$ where $i\neq l,\,i\in \mathcal{I}$ to the wireless link $l$ between the UE and the UAV-BS can be calculated as follows,
\begin{equation}
    I_{\textrm{mW}}^{i,l}=f_{\textrm{dBm-mW}}\left(\mathcal{G}_{\textrm{dBi}}^{i,\textrm{Tx}}\left(\phi^{\dagger},\theta^{\dagger}\right)+\mathcal{P}_{\textrm{dBm}}^{i,\textrm{Tx}}-L\left(d_{\left(i,l\right)}\right)\right),
\label{eq:interference}
\end{equation}
where $\mathcal{P}_{\textrm{dBm}}^{i,\,\textrm{Tx}}$ is the transmit power used for the transmission at the wireless link $i$. $L\left(d_{\left(i,l\right)}\right)$ is the signal attenuation to the path loss, as formulated in~\eqref{eq:Loss}.
The magnitude of path loss depends on $d_{\left(i,l\right)}$, which is the distance between the $i$-th transmitter and the $l$-th receiver. Moreover, $\mathcal{G}_{\textrm{dBi}}^{i,\,\textrm{Tx}}$ is an antenna radiation gain between the $i$-th transmitter and the $l$-th receiver depending on $\phi$ and $\theta$, which are the angular differences between these respective transmitter and receiver, as formulated in~\eqref{eq:radiation}. Lastly, $f_{\textrm{dBm-mW}}$ is a function for unit translation from dBm to milli watt (mW) scale, where $f_{\textrm{dBm-mW}}\triangleq10^{\left(x/10\right)}$.

\subsubsection{Capacity Calculation With Shannon's Formula}

This paper assumes Shannon's formula to calculate the theoretical data rates of 60\,GHz mmWave wireless systems. The Shannon's formula gives the channel capacity $\mathcal{C}_{l}\left(d\right)$ in~\eqref{eq:Shannon's} at a wireless link $l$ between a transmitter and a receiver separated by a distance $d$, as follows,
\begin{equation}\label{eq:Shannon's}
\mathcal{C}_{l}\left(d\right)=\textbf{\textrm{W}}\cdot\log_{2}\left(1+\frac{\mathcal{P}_{\textrm{mW}}^{l,\,\textrm{Rx}}\left(d\right)}{n_{\mathrm{mW}}+\sum_{i\in\mathcal{I},\,i\neq l}I_{\textrm{mW}}^{i,l}}\right),
\end{equation}
where \textbf{\textrm{W}} is a 60\,GHz channel bandwidth ($2.16\,\textrm{GHz}$ in the case of IEEE 802.11ad~\cite{kim2017feasibility}). Here, $\mathcal{P}_{\mathrm{mW}}^{l,\mathrm{Rx}}\left(d\right)$ is the received signal power in a mW scale at the wireless link of the $l$-th receiver, $I_{\textrm{mW}}^{i,l}$ is the interference in~\eqref{eq:interference}, and $n_{\mathrm{mW}}$ is the background noise in a mW scale. Note that $\mathcal{I}$ is the set of all given wireless links.
\begin{itemize}
    \item $\mathcal{P}_{\mathrm{dBi}}^{l,\mathrm{Rx}}\left(d\right)$ is calculated as follows,
    \begin{equation}
            \mathcal{P}_{\mathrm{dBi}}^{l,\mathrm{Rx}}\left(d\right)=\underbrace{\mathcal{G}_{\mathrm{dBi}}^{l,\mathrm{Tx}}+\mathcal{P}^{l,\mathrm{Tx}}_{\mathrm{dBm}}}_{\textrm{EIRP}}-L\left(d\right)+\mathcal{G}_{\mathrm{dBi}}^{l,\mathrm{Rx}},
    \label{eq:RSS}
    \end{equation}
    where $\mathcal{G}_{\mathrm{dBi}}^{l,\mathrm{Tx}}$ and $\mathcal{P}^{l,\mathrm{Tx}}_{\mathrm{dBm}}$ are a transmit antenna gain and a transmit power at the Tx of wireless link $l$. $L\left(d\right)$ is a path loss when Tx and Rx are separated by distance $d$ in~\eqref{eq:Loss}. Lastly, $\mathcal{G}_{\mathrm{dBi}}^{l,\mathrm{Rx}}$ is a receive antenna gain ($3\,\mathrm{dBi}$ in~\cite{kim2017feasibility}). As defined in~\cite{kim2017feasibility}, the United States limits the maximum value of equivalent isotropic radiated power (EIRP) in the $60$\,GHz mmWave radio channel to $43$\,dBm~\cite{6732923}, \textit{i.e.,} the summation of $\mathcal{G}_{\mathrm{dBi}}^{l,\mathrm{Tx}}$ and $\mathcal{P}^{l,\mathrm{Tx}}_{\mathrm{dBm}}$ cannot exceed $43$\,dBm.
    Under the consideration of the RF hardware and parameter settings for an existing $60$\,GHz antenna system~\cite{kim2017feasibility}, $\mathcal{G}_{\mathrm{dBi}}^{l,\mathrm{Tx}}=19\,\mathrm{dBi}$ (transmit antenna gain in a dBi scale) and $\mathcal{P}_{\mathrm{dBm}}^{l,Tx}=24\,\mathrm{dBm}$ (transmit power in a dBm scale at $l$).
    \item $n_{\mathrm{mW}}$ is the background noise and calculated as follows,
    \begin{equation}\label{eq:noise}
        n_{\textrm{mW}}=f_{\textrm{dBm-mW}}\left(k_{\textrm{B}}T_{e}+10\log_{10}\textbf{\textrm{W}}+\sigma\right),
    \end{equation}
    where $k_{\textrm{B}}T_{e}$ is the noise power spectral density ($-174\textrm{dBm/Hz}$ in~\cite{kim2017feasibility}) and $\sigma$ is the system-specific additional loss, which is the summation of implementation loss ($10\,\textrm{dB}$ in~\cite{kim2017feasibility}) and noise figure ($5\,\textrm{dB}$ in~\cite{kim2017feasibility}).
\end{itemize}

\subsubsection{Actual Data Rate Computation with IEEE 802.11ad MCS}\label{sec:MCS}
The theoretical data rate by Shannon's capacity formula served by UAV-BSs is expressed in~\eqref{eq:Shannon's} and this is the fundamental limit of the wireless communications. However, due to hardware and RF implementation limitations, actual data rates will be lower than the theoretical formulation. Therefore, in this paper we calculate the actual data rate using the MCS table defined in IEEE 802.11ad~\cite{kim2016performance}. First of all, the Rx signal strength under the consideration of interference components is calculated. After that, the actual maximum supportable data rate $A_l$ of the 60\,GHz radio link $l$ served by the UAV-BS is obtained by Table~\ref{tab:MCS}.

\subsubsection{The Use of 60\,GHz Wireless Communications for Mobile Access}
Our considered mmWave (IEEE 802.11ad based) wireless system has a large channel bandwidth, low-latency transmission, high beam directivity, high diffraction, and high scattering~\cite{niu2015survey}.
These characteristics of mmWave wireless systems have the advantage of being less sensitive to interference from nearby mobile access (\textit{i.e.,} spatial reuse) while being very sensitive to blocking~\cite{niu2015survey}.
In particular, achieving ultra-low latency in the V2X communication among autonomous vehicles is crucial, as it significantly reduces the driving distance before receiving information~\cite{9447828, 8080373, 7355568}. This is especially important due to the dynamic and time-varying nature of the environments in which autonomous vehicles operate.
In an urban environment, for example, blocking by high-rise buildings may affect mmWave wireless systems. However, our proposed algorithm can minimize this problem through the optimal positioning of UAV-BSs cooperatively. In that case, it is possible to provide a higher QoS service to dense UEs using high directivity and low latency compared to the existing generation of mobile networks. 
In rural and suburban environments, there are fewer problems with blocking, and the advantages of mmWave can be further maximized~\cite{5733382}.

\begin{table}[t]
\centering
\caption{60\,GHz IEEE 802.11ad MCS Table~\cite{kim2017feasibility}.}
\renewcommand{\arraystretch}{1.3}
\begin{tabular}{c||r|r|r}
\toprule[1pt]   
\textbf{Rx} & \textbf{Available} & \textbf{Maximum} & \textbf{Capacity Estimation} \\
\textbf{Sensitivity} & \textbf{MCS} & \textbf{Data Rates} & \textbf{with Shannon~\eqref{eq:Shannon's}} \\ \midrule
-78\,dBm & MCS0 & 27.5\,Mbps & 1.43\,Gbps \\
-68\,dBm & MCS1 & 385\,Mbps & 2.04\,Gbps \\
-66\,dBm & MCS2 & 770\,Mbps & 2.40\,Gbps \\
-65\,dBm & MCS3 & 962.5\,Mbps & 2.81\,Gbps \\
-64\,dBm & MCS4 & 1155\,Mbps & 3.25\,Gbps \\
-63\,dBm & MCS6 & 1540\,Mbps & 3.74\,Gbps \\
-62\,dBm & MCS7 & 1925\,Mbps & 4.25\,Gbps \\
-61\,dBm & MCS8 & 2310\,Mbps & 5.38\,Gbps \\
-59\,dBm & MCS9 & 2502.5\,Mbps & 7.90\,Gbps \\
-55\,dBm & MCS10 & 3080\,Mbps & 8.57\,Gbps \\
-54\,dBm & MCS11 & 3850\,Mbps & 9.23\,Gbps \\
-53\,dBm & MCS12 & 4620\,Mbps & 43.48\,Gbps \\
\bottomrule[1pt]
\end{tabular}
\label{tab:MCS}
\end{table}

\section{CTDE-based MADRL for Autonomous Multiple UAV-BS Positioning}\label{sec:method}
This section introduces our proposed architecture for autonomous UAV-BS positioning to increase the UAV-BS coverage in order to guarantee a certain level of connectivity and stability. First of all, we explain the proposed system settings and assumptions (refer to Sec.~\ref{sec:system-setting-and-assumptions}), and then the details of our proposed algorithm are described (refer to Sec.~\ref{sec:details}).

\subsection{System Settings and Assumptions}\label{sec:system-setting-and-assumptions}

Our proposed autonomous UAV-assisted network consists of $N$ UEs, $M$ agent UAV-BSs, and $K$ non-agent UAV-BSs \tblue{as illustrated in Fig.~\ref{fig:system model}. There are one leader UAV-BS and $M-1$ non-leader agent UAV-BSs among agent UAV-BSs. Due to variations in UE density between densely and sparsely populated areas, the demand for services may vary, making it necessary to develop an aerial mobile access network model that takes the number of UE into account.}
We denote the sets of UEs, agent UAV-BSs, and non-agent UAV-BSs as
$\mathcal{U} \triangleq \left\{ u_{1}, u_{2}, \dots, u_n, \dots, u_{N} \right\}$, 
$\mathcal{B} \triangleq \left\{ b_{1}, b_{2}, \dots, b_m, \dots, b_{M} \right\}$, 
$\mathcal{H} \triangleq \left\{ h_{1}, h_{2}, \dots, h_k, \dots, h_{K} \right\}$, respectively.
Hereafter, we use indicator variables $u_n$ and $b_m$ denoted as $n$-th UE and $m$-th agent UAV-BS where $\forall u_n \in \mathcal{U}, n \in [1, N]$, $\forall b_m \in \mathcal{B}, m \in [1, M)$. Note that $M$-th UAV-BS is a leader UAV-BS, and it is denoted as $b_M$.

\tblue{A leader agent UAV-BS is to handle communications between non-leader agent UAV-BSs. 
The leader agent UAV-BS receives information for multi-agent cooperation as a role of communication controller at a higher altitude than other non-leader agents, as depicted in Fig.~\ref{fig:control system}. That is, the leader UAV-BS uses the information from all the UAV-BSs, while non-leader agent partially observes the information. On the other hand, the non-agent UAV-BS has no observations and does not participate in the UAV cooperation.}
In the uplink transmission of UAV-BSs, the cooperative UAV-BSs transmit their observation information to the controller. 
This paper assumes that each UE is associated with only one UAV, whereas each UAV-BS can support multiple UEs. 
The positions of UEs and UAV-BSs are represented by ($x$, $y$, $z$) in the 3D grid. The 3D positions of UEs are randomly distributed, and also time-varying.

As mentioned in Sec.~\ref{sec:3-b}, we assume that all UAV-BSs provide a 60\,GHz mmWave spectrum for higher directivity and data rate. Their coverage radius can be calculated as follows,
\begin{eqnarray}
    C(b_{m},u_{n}) = \frac{H(b_{m})-H(u_{n})}{H(b_{m})} \times \tan\left(\theta_W/2\right),
    \label{eq:altitude_Agent}\\
    C(h_{k},u_{n}) = \frac{H(h_{k})-H(u_{n})}{H(h_{k})} \times \tan\left(\theta_W/2\right),
    \label{eq:altitude_nonAgent}
\end{eqnarray}
where $C(\cdot)$ and $H(\cdot)$ return the coverage areas of $m$-th agent UAV-BS and $k$-th non-agent UAV-BS from the perspective of the $n$-th UE and the altitude of the input, respectively. $\theta_W$ is the UAV-BS's directional antenna beamwidth, which is set to $80^\circ$ in Table~\ref{tab:setup}.
Since the altitude of UEs is different, we must consider each UE's elevation when calculating the coverage radius. According to \eqref{eq:altitude_Agent} and \eqref{eq:altitude_nonAgent}, the coverage radius becomes smaller when the UE's altitude becomes higher or UAV-BS's altitude becomes lower.
We also assume that UEs can request one type of communication service from among HD video streaming, online gaming, web surfing, and voice-over IP. Here, HD video streaming is the service that requires the highest data rate, which is up to 4\,Gbps~\cite{8642333}. Once a UE is within the coverage radius of any UAV-BS, the UE can obtain sufficient data rates for HD video streaming thanks to high directivity and data rates of 60\,GHz mmWave.

However, the data rate is not an absolute value of the traffic. UEs do not always receive their services with acceptable quality. For example, packet-based networks carry large amounts of traffic. The network must forward the traffic under the consideration of any application's performance requirements. To achieve the requirements of UEs, the network devices (\textit{e.g.,} router, switch) forward traffic considering QoS. 
It allows network devices to apply different behaviors to traffic after distinguishing the type of service. This paper define a quality function
that consists of two components of QoS, \textit{i.e.,} \textit{i)} video streaming and \textit{ii)} the others~\cite{jung2021infrastructure}, where the QoS component increases in a non-linear form using a sigmoid function for video streaming whereas it increases in a logarithmic-like function as the data rate $A_l$ increases for the others. Note that $A_{l}$ is the actual maximum supportable data rate calculated by matching capacity estimation with Shannon and MCS Table in~\ref{sec:MCS}. Finally, the quality function can be as follows,

\begin{equation}
    f\left(A_l\right)=
    \begin{cases}
        \left(1+\exp^{-v_{a}\left(A_l-w_{a}\right)}\right)^{-1},\;\textit{(video traffic)},\\
        \log\left(v_{b}\cdot A_l +w_{b}\right),\;\textit{(otherwise)},
    \end{cases}
\label{eq:quality}
\end{equation}
where the values of weight parameters are $v_a$\,=\,0.01, $w_a$\,=\,1024, $v_b$\,=\,1, and $w_b$\,=\,1, respectively~\cite{jung2021infrastructure}.

For the autonomous UAV-BSs' cooperation, this paper assumes that the link between UAVs and the leader UAV-BS is good enough to deliver UAV-BS information without losing robust and stable MADRL training. It is justifiable because the code rate used in leader UAV-BS and UAV-BSs communications is small enough. 
However, the link between UAV-BSs is limited.
Lastly, each UAV has a maximum battery capacity in the beginning.
The proposed autonomous coordination system aims to increase the number of UEs serviced by UAV-BSs under certain amounts of service quality guarantees in an uncertain environment. In order to take care of the uncertainty, we consider the number of UEs which can be varied due to each UAV's conditions, \textit{e.g.,} the possibility for UAVs drop, malfunction, or energy-exhaust.

\subsection{Coordinated MADRL-based Algorithm Design Details}\label{sec:details}

\subsubsection{Reinforcement Learning Formulation}
Our considering UAV-BS positioning problem is formulated with Markov decision process (MDP). It can be denoted as $\left(\mathcal{S}, \mathcal{A}, \mathcal{R}, \mathcal{P}, \gamma\right)$ which are state space, action space, reward space, transition probability, and discount factor, respectively. 

\BfPara{State} The state space in our problem consists of three parts, \textit{i.e.,} location information, energy information, and service connection information. More details are as follows.

\begin{itemize}
\item \textit{Location Information:} Each UAV-BS has to observe its own location information (\textit{e.g.,} position information and distance information with UEs or other UAV-BSs) to identify whether each UE is serviced by a UAV-BS or not, avoiding collisions with other UAV-BSs. The positions of each UAV-BS, UE and non-agent UAV-BS are defined as $p(b_{m})$, $\forall m \in [1, M]$, $p(u_{n})$, $\forall n \in [1, N],$ and $p(h_{k}), k \in [1, K]$, respectively. Note that the positions of all UAV-BSs and UEs are represented using their own $x$-, $y$-, and $z$-coordination values, \textit{i.e.,} 3D geometry. The distances for $b_{m}$ with $b_{m}^{′}$, $u_{n}$, and $h_{k}$ are denoted as $d(b_{m}, b_{m}^\prime$), $d(b_{m}, u_{n}),$ and $d(b_{m}, h_{k})$, where $d(b_{m}, (\cdot))$ means a physical distance between $b_{m}$ and $(\cdot)$. Therefore, the location information of $b_{m}$ is $p_m\triangleq \{p(b_{m}), d(b_{m}, x)\}$, where $x \in \{B\cup U\cup H\}$\textbackslash $b_{m}$.

\item \textit{Energy Information:} Each agent UAV-BS has to observe its energy information to prevent full discharging. If any UAV-BS is fully discharged, they cannot provide reliable communication services to their associated UEs. UAV-BSs consume operational energy for aviation and wireless communications. However, this paper considers only energy consumption by aviation because communication-related energy is usually much lower than aviation energy~\cite{tvt202106jung}. Notice that communication-related energy consumption is only a few watts, whereas aviation-related energy consumption is a few hundred watts. We denote the amount of energy remaining of $b_{m}$ as $e_{m}$. For the initial status of operations, we assume that each agent UAV-BS is fully charged, and we adopt the UAV queue backlog model as an energy discharge model, as shown in~\eqref{eq:Q}~\cite{jung2020joint}. As a result, $e_m$ at $t$ can be represented as following $Q_m(t)$, \textit{i.e.,} 
\begin{equation}\label{eq:Q}
    Q_{m}(t+1)=\textrm{max}\,\{0, \,Q_{m}(t)-\mu_m(t)\}.
\end{equation}
where $\mu_m(t)$ is the energy expenditure of $b_{m}$ at $t$. There are two operational modes related to discharging energy at UAV-BS, \textit{i.e.,} hovering and round-trip traveling~\cite{jung2020joint}. 
The energy expenditure modeling formulation of $m$-th UAV-BS at $t$ for hovering can be expressed as~\eqref{eq:hovering}~\cite{zeng2017energy}, 
\begin{equation}\label{eq:hovering}
    \mu_m(t) = 
        \underbrace{\frac{\delta}{8}\rho sA\Omega^3R^3}_{\textit{bladeprofile}, \,P_{o}}
        +\underbrace{\left(1+k\right)\frac{W^{3/2}}{\sqrt{2\rho A}}}_{\textit{induced}, \,P_{i}},\qquad \textit{(hovering)},
\end{equation}
where $\delta$, $\rho$, $s$, $A$, $\Omega$, $R$, $k$, and $W$ are the profile drag coefficient, air density, rotor solidity, rotor disc area, blade angular velocity, rotor radius, an incremental correction factor to induced power, and aircraft weight including battery and propellers, respectively. 
In addition, the energy expenditure modeling formulation of $m$-th UAV-BS at $t$ for round-trip traveling can be expressed as following~\eqref{eq:roundtrip}~\cite{zeng2017energy}, 
\begin{equation}\label{eq:roundtrip}
\begin{split}
    \mu_m(t) = 
        \underbrace{P_{o}\left(1+\frac{3v^{2}}{U_{tip}}\right)}_{\textit{bladeprofile}}
        +\underbrace{P_{i}\left(\sqrt{1+\frac{v^{4}}{4v_{0}^{4}}}-\frac{v^{2}}{2v_{0}^{2}}\right)^{0.5}}_{\textit{induced}}\\
        +\underbrace{\frac{1}{2}d_{0}\rho sAv^{3}}_{\textit{parasite}},\qquad\textit{(round-trip traveling)},
\end{split}
\end{equation}
where $P_o$ and $P_i$ are the blade profile power, and induced power in~\eqref{eq:hovering}, respectively. $v$, $U_{tip}$, $v_0$, and $d_0$ are the flight speed at time $t$, tip speed of the rotor blade, mean rotor-induced velocity in hovering, and fuselage drag ratio, respectively. The values of these parameters in~\eqref{eq:hovering} and~\eqref{eq:roundtrip} are summarized in Table~\ref{tab:parameters of uav}.
As~\eqref{eq:Q}, the remaining energy of the UAV-BS at $t$ and energy expenditure $\mu_m(t)$ determines the remaining energy of the UAV at $t+1$. If the value of $Q_m(t)-\mu_m(t)$ is negative, the UAV has no energy left. Therefore, the UAV-BS has zero residual energy when the time is $t+1$.

\item \textit{Service Information:} This information is needed to be observed by each UAV-BS to identify whether UE obtains wireless communication services or not. This paper defines service information as $c_{nm}$ (service availability from $m$-th agent UAV-BS to its associated $n$-th UE) and $c_{nk}$ (service availability from $k$-th non-agent UAV-BS to its associated $n$-th UE), which are binary, \textit{i.e.,} $c_{nm}\in\{0,1\}$ and $c_{nk}\in\{0,1\}$. If the $n$-th UE obtains service by $m$-th or $k$-th UAV-BS, the corresponding $c_{nm}$ or $c_{nk}$ will be $1$ (otherwise, $0$). In addition, it is true that each UAV-BS can service at most one UE, \textit{i.e.,} $c_{nm}$ and $c_{nk}$ cannot have $1$ at the same time. Accordingly, the energy information of $b_m$ is $c_m\triangleq\{c_{nm}, c_{nk}\}$.
\end{itemize}

Finally, the set of states can be represented as $S\triangleq \{s_1, \dots, s_m, \dots, s_M\}$ where $s_m$ represents the state of $m$-th UAV-BS, defined as $s_m\triangleq \{p_m, e_m, c_m\}$. 
 In this paper, we also consider the observation information $o_m$ for each agent UAV-BS. Note that the observation $o_m$ is a partial information of $s_m$. We will elaborate on the observation next.
 
\begin{table}[t]
\centering
\caption{Actual Specification of UAV~\cite{zeng2019energy}.}
\renewcommand{\arraystretch}{1.0}
\begin{tabular}{l||r}
\toprule[1pt]
\textbf{\textsf{Notation}} & \textbf{\textsf{Value}} \\ \midrule
Flight speed, $v$ & 20 m/s \\
Average maximum flight time & 30 min \\
Capacity of flight battery & 5,870 mAh\\
Voltage & 15.2 V\\
Aircraft weight including battery and propellers, $W$ & 1375 g\\ Rotor radius, $R$ & 0.4 m \\
Rotor disc area, $A=\pi R^{2}$ & 0.503 $m^{2}$ \\
Number of blades , $b$ & 4 \\
Rotor solidity, $s, \frac{0.0157b}{\pi R}$ & 0.05 \\
Blade angular velocity, $\Omega$ & 300 radius/s\\
Tip speed of the rotor blade , $U_{tip}=\Omega R^{2}$ & 120 \\
Fuselage drag ratio, $d_{0}=\frac{0.0151}{sA}$ & 0.6 \\
Air density, $\rho$ & 1.225 kg/$m^{3}$ \\
Mean rotor-induced velocity in hovering, $v_{0}=\sqrt\frac{W}{s\rho A}$ & 4.03 \\
Profile drag coefficient, $\delta$ & 0.012 \\
Incremental correction factor to induced power, $k$ & 0.1 \\
Maximum service ceiling & 6000 m \\
\bottomrule[1pt]
\end{tabular}
\label{tab:parameters of uav}
\end{table}

\BfPara{Action} The action space consists of 7 discrete actions to move. 
At every time step, agent UAV-BSs can move $\left(v \times t\right)$ in $x$ or $y$ or $z$ direction or hover in place, where $v$ and $t$ denote the speed of the UAV-BS, and unit-time. Each UAV-BS makes action decision from the discrete set of actions, \textit{i.e.,} $\mathcal{A} = \{hovering, x_{pos} \pm \left(v \times t\right), y_{pos} \pm \left(v \times t\right), z_{pos} \pm \left(v \times t\right)\}$. 

From UAV-BSs' action decision, the proposed system can achieve wireless communication service as many UEs as possible in high quality. 
To be specific, agent UAV-BSs have three $z$-coordinates, where $\forall z_{pos} \in \left[1500\,\text{m}, 2000\,\text{m}, 2500\,\text{m}\right]$, which is justifiable according to \cite{mozaffari2016efficient}.
In~\eqref{eq:altitude_Agent} and ~\eqref{eq:altitude_nonAgent}, the coverage radius increases is proportional to the altitude. In addition, the path loss (referred to \eqref{eq:Loss}) increases as the distance between the UE and the UAV-BS $d$ increases. 
As the path loss increases, the QoS eventually decreases according to \eqref{eq:RSS},~\eqref{eq:Shannon's}, \eqref{eq:quality}, and Table~\ref{tab:MCS}. There is a trade-off between a UAV's coverage radius and the UE's QoS. In summary, the optimal action can maximize the QoS and the number of supporting UEs.

\BfPara{Reward} Our proposed method has two types of rewards, \textit{i.e.,} \textit{i)} individual and \textit{ii)} common rewards. The individual reward is for preventing agent UAV-BSs from fully discharging and also for providing high-quality services. In addition, the common reward is for conducting agent UAV-BSs cooperation.
\begin{itemize}
\item \textit{Individual Reward:} The individual reward is composed of energy reward $r_m^e$ and quality reward $r_m^u$. First, energy reward $r_m^e$ is related to energy information $e_m$ for stable operation of UAV-BS using the UAV queue backlog model representing discharge of UAV-BS in~\eqref{eq:Q}. 
Next, the quality reward $r_m^u$ is defined as the total sum of UEs' QoS serviced by the agent UAV-BS for reliable service. 
The energy reward $r_m^e$ and quality reward $r_m^u$ are as,
\begin{eqnarray}
    r_m^e &=& \textrm{max}\,(0, \;e_{m}),\label{eq:energy} \\
    r_m^u &=& \sum^N_{n=1}\nolimits f(A_{l}^{\chi,n}), \forall \chi \in \{m, k \}\label{eq:quality reward}
\end{eqnarray}
where $f(\cdot)$ returns the quality of the input, which depends on a service type (\textit{e.g.,} video traffic or the others). In addition, $A_l^{\chi,n}$ stands for $n$-th UE's achievable data rate of the wireless link between $m$-th or $k$-th UAV-BS and $n$-th UE,$ \forall \chi \in \{m, k \}$, which is calculated in Sec.~\ref{sec:MCS}. If UEs are located within the coverage of multiple UAV-BSs, they will be served by one UAV-BS which provides the most robust link.

\textit{Common Reward:}
The common reward represents the overall system reliability, and note that all agent UAV-BSs receive the same common reward at the same time. The two equations below represent the components of the common reward.
\begin{equation}
    \tau = \frac{1}{N} \sum_{n=1}^N\nolimits \sum_{m=1}^M \nolimits\sum_{k=1}^K\nolimits (c_{nm} + c_{nk}), \label{eq:support rate}
\end{equation}
where $c_{nm}$ and $c_{nk}$ are binary variables, \textit{i.e.,} $c_{nm}\in\{0,1\}$ and $c_{nk}\in\{0,1\}$. Note that $c_{nm}=1$ when $m$-th agent UAV-BS provide mmWave access services to $n$-th UE (and $c_{nm}=0$ for others). Similarly, $c_{nk}=1$ when $k$-th non-agent UAV-BS provide service to $n$-th UE. Accordingly, the summation of $c_{nm}$ and $c_{nk}$ has the number of UEs as the maximum value and zero as the minimum value. By averaging the summation of $c_{nm}$ and $c_{nk}$, the UE support rate $\tau$ can be obtained.

In order to improve the quality of the service provided to users, it is necessary to reduce $\omega$ to decrease interference from other UAV-BSs in \eqref{eq:interference}, and \eqref{eq:Shannon's}. Therefore, we define the overlapping degree of the UAV-BS service area as, 
\begin{equation}
    \omega = 1-\frac{\{\cap_{m=1}^{M}S\left(R_{m}\right)\}\bigcap\{\cap_{k=1}^{K}S\left(R_{k}\right)\}}
    {\{\cup_{m=1}^{M}S\left(R_{m}\right)\}\bigcup\{\cup_{k=1}^{K}S\left(R_{k}\right)\}}, \label{eq:overlapped}
\end{equation}
where $S(\cdot)$ returns the area of the input which corresponds to the coverage area of the $m$-th agent UAV-BS or the $k$-th agent UAV-BS.
For providing reliable mmWave access service to UEs, agent UAV-BSs should increase the support rate $\tau$, while decreasing the overlapping degree $\omega$. Thus, the total reward can be finally expressed as follows, 
\begin{equation}\label{eq:common reward}
    r^c = \frac{\tau}{1+\omega},
\end{equation}
and thus, the common reward increases, as the support rate $\tau$ increases as well as the degree of overlap $\omega$ decreases. 
Finally, the total reward for each agent UAV-BS $r_m^{total}$ can be represented follows,
\begin{equation}\label{eq:total reward}
    r_m^{total} = \underbrace{r_m^e \times r_m^u}_{\textrm{Individual rewards}} \times \underbrace{r^c}_{\textrm{Common reward}}.
\end{equation}
\end{itemize}

\begin{figure*}[t!]
    \centering
    \includegraphics[width=1.8\columnwidth]{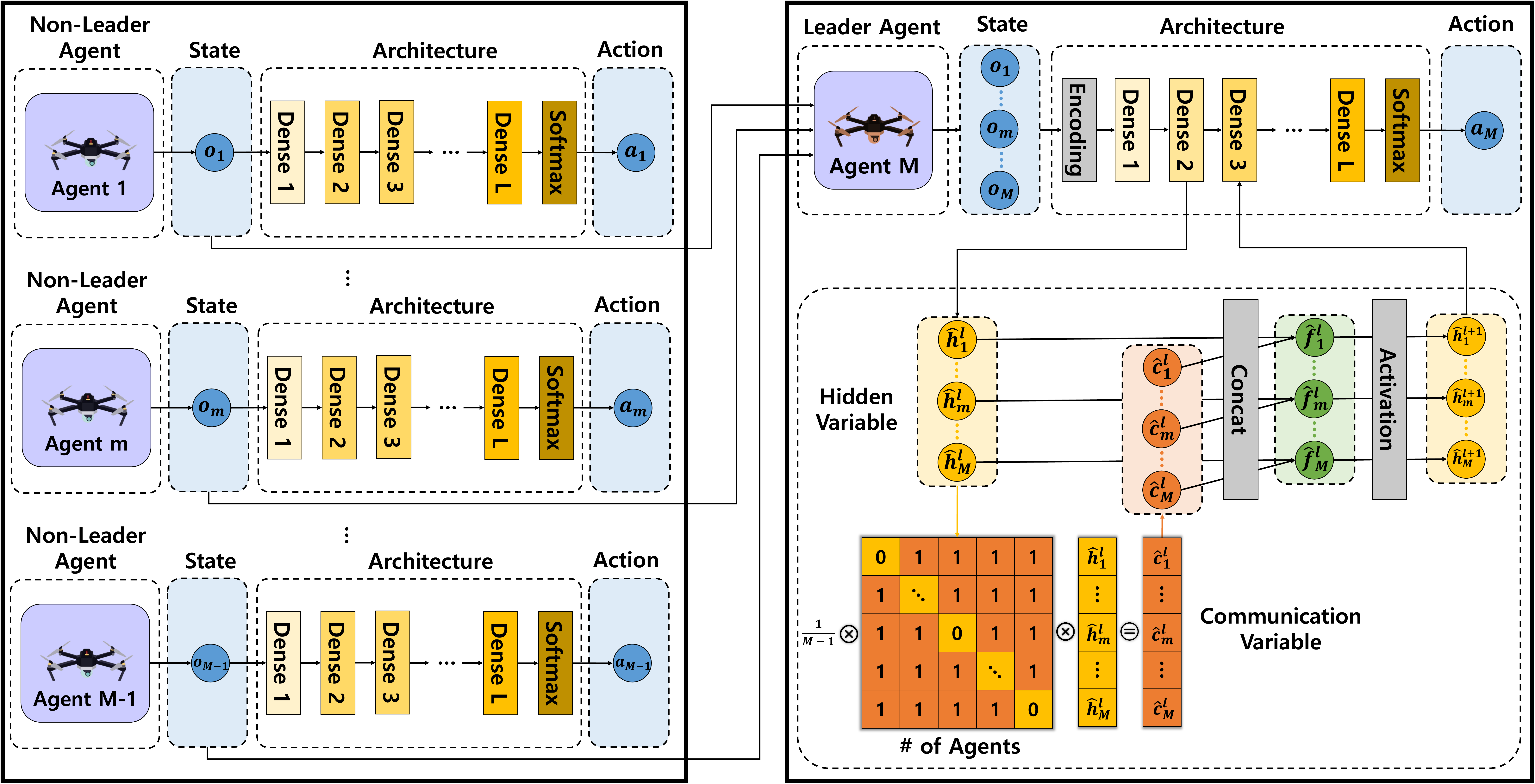}
    \caption{Coordination system of the CommNet and DNN-based policy.}
    \label{fig:control system}
\end{figure*}
\subsubsection{Information Sharing via Inter-Agent Communication}
To cooperate with UAV-BSs during MADRL training, we use CommNet architecture that the communication between a leader UAV-BS and UAV-BSs is available.
 
 First of all, the leader UAV-BS collects and distributes other UAV-BSs' observation information~\cite{sukhbaatar2016learning}. For mathematical amenability, we set the input state as a set of observation in matrix form, \textit{i.e.,} $s \equiv [o_1,\cdots ,o_m,o_{M-1}]$. As shown in Fig.~\ref{fig:control system}, a leader-UAV BS makes state $s$ by collecting other agent UAV-BSs' observations. The state is an input of CommNet. 
The CommNet architecture follows the process below, 
\begin{align}
    \hat{h}^{1}_{m} =& \textsf{Encoder}(o_m),  
    \label{eq:encoding}\\
    \hat{c}^{l}_{m} =& \frac{1}{M-1} \sum_{m \neq M}\nolimits \hat h^{l}_{m}, 
    \label{eq:comm}\\
    \hat{h}^{l+1}_{m} =& \textsf{Activ}(\hat{f}^{l}_{M}(\textsf{Concat}(\hat{h}^{l}_{m}, \hat{c}^{l}_{m}))), 
    \label{eq:hidden}\\
    {a}_M =& \textsf{Softmax}(\textsf{Concat}([\hat{h}^{L+1}_1,\cdots,\hat{h}^{L+1}_M])), 
    \label{eq:hidden2}
\end{align}
where $\textsf{Activ}(\cdot)$, $\textsf{Concat}(\cdot)$, $\textsf{Encoder}(\cdot)$, $\hat{f}^{l}_{M}$, and $\textsf{Softmax}(\cdot)$  stand for an activation function (\textit{e.g.,} ReLU, sigmoid, or tangent-hyperbolic), concatenate function, encoding function, the parametrized non-linear function of layer $l$, and softmax function, respectively. 
Note that $\hat{\cdot}$ is to distinguish whether the hidden/communicate variables and layers are based on CommNet.  \eqref{eq:encoding} describes that every row of state (\textit{e.g.,} observation $o_m$) is encoded to hidden variable $\hat h^1_m$. \eqref{eq:comm} depicts that the communication variable $ \hat c^l_m$ is obtained by averaging hidden variable of agents except $m$-th UAV-BS's for every layer $l$. The information exchange between inter-agents occurs by averaging hidden variables.  
Next, \eqref{eq:hidden} describes the variables $\{ \hat h^l_m, \hat c^l_m\}$ are feed-forwarded to the next layer. 
In CommNet, the processes \eqref{eq:comm} and \eqref{eq:hidden} is repeatedly conducted for $l \in [1,L]$. The leader UAV-BS's action probability $\pi_M(a|s)$ which are made by the sequential processes of concatenating all hidden variables and taking a softmax function. 
Second, the agent UAV-BS makes an action decision with DNN-based policy, which can be expressed as,
\begin{equation}
    {h}^{l+1}_{m} =\textsf{Activ}({f}^{l}_{m}({h}^{l}_{m})).
\end{equation} 
The input of first layer is an observation, \textit{i.e.,} ${h}^{l}_{m} = o_m$. Then, the action distribution $\pi_m(a_m|o_m)$ is obtained by taking the softmax function to $h^L_m$. 
Note that the communication between inter-UAV is not related to agent UAV-BS's action decision.
The CommNet-based leader UAV-BS employs the method of information sharing by transmitting encoded hidden variables instead of directly relaying the observed information. This approach enhances the security and privacy of the communication process similar to federated learning (FL)~\cite{baek2022joint}. Consequently, in sensitive autonomous driving scenarios such as ambulances operating within hospital settings where secure information sharing is crucial, the utilization of CommNet-based mutual communication ensures the secure exchange of sensitive information among agent UAV-BSs while preserving data privacy.

\begin{algorithm}[t]
\small
    Initialize parameters of \textit{centralized critic} network $\boldsymbol{\theta}^{Q}$ and multiple \textit{actor} networks $\boldsymbol{\theta}^{\mu}_m$, $\forall m \in [1,M]$\\
    Update parameters of target networks as: 
    $\boldsymbol{\theta}^{\hat{Q}} \leftarrow \boldsymbol{\theta}^{Q}$,
    $\boldsymbol{\theta}^{\hat{\mu}}_m \leftarrow \boldsymbol{\theta}^{\mu}_m$, 
    $\forall m \in [1,M]$ \\
    \For{episode = 1, MaxEpisode}{
            $\triangleright$ Initialize \textbf{Mobile Access Environments}\\
            \For{time step = 1, T}{
            \For{each UAV-BS $m$}{
                $\triangleright$ Select the action based on its policy $\pi_m(a_m|o_m;\boldsymbol{\theta}^{\mu})$ at time step $t$ \\
            }
            $\triangleright$ Get reward ${R}_{total}$ and next set of states ${S}'$ by executing actions in \textbf{Simulation Environments}\\
            $\triangleright$ Store the transition pairs $\xi = ({S}, {A}, {R}_{total}, {S}')$ in the replay buffer $\Phi$\\
            \textbf{If} \textit{time step} \textbf{is update period and}  \textit{replay buffer} \textbf{is full enough to train, do followings:}\\
            $\triangleright$ Sample a mini-batch randomly from $\Phi$\\
            $\triangleright$ Set $y_{i} = r_{i} + \gamma \hat{Q}(s_{i}^{'}, \pi(s_{i}^{'}|\boldsymbol{\theta}^{\hat{\mu}})|\boldsymbol{\theta}^{\hat{Q}})$ \\
            $\triangleright$ Update $\boldsymbol{\theta}^{Q}$ by stochastic gradient descent to the loss function of \textit{centralized critic} network: $L = \frac{1}{\varphi}\sum_{i}{(y_{i} - Q(s_{i}, a_{i}|\boldsymbol{\theta}^{Q}))}^{2}$ \\
            \For{each UAV-BS $m$}{
            $\triangleright$ Update $\boldsymbol{\theta}^{\mu}_m$ by stochastic gradient ascent with respect to the gradient of objective function in \textit{actor} networks: \\
                $\nabla_{\boldsymbol{\theta}}J(\boldsymbol{\theta}^{\mu}_{m}) \approx \mathbbm{E}[{Q(s, a|\boldsymbol{\theta}^{Q})\nabla_{\boldsymbol{\theta}}\pi(s|\boldsymbol{\theta}^{\mu})}]$ \\
            }
            \textbf{If} \textit{episode} \textbf{reaches at target network update cycle, do followings:}\\
            $\triangleright$ Update parameters of target networks as: $\boldsymbol{\theta}^{\hat{Q}} \leftarrow \boldsymbol{\theta}^{Q}$, $\boldsymbol{\theta}^{\hat{\mu}}_m \leftarrow \boldsymbol{\theta}^{\mu}_m$, $\forall m \in [1,M]$
            }
        }
    \caption{Overall training process for instructing multi-UAV to collaboratively operate an autonomous mobile access network with the goal of providing high-performance wireless communication services.}
    \label{alg:CommNet}
\end{algorithm}

Next, we describe the overall CTDE training procedure in our proposed coordination framework. We utilizes multi-agent actor-critic DRL framework \cite{lowe2017multi}. We denote the agents have \textit{actor} policy of which parameters are denoted as $\boldsymbol{\theta}^\mu_m$, $\forall m \in [1, M]$. The \textit{centralized critic} is used where the critic parameters are denoted as $\boldsymbol{\theta}^Q$. 

\subsection{Summary and Algorithm Pseudo-Code}
\tblue{In summary, this entire computational procedure is illustrated in Fig.~\ref{fig:control system}. This illustration is for the training system which is composed of the CommNet-based policy for leader UAV-BS and the $M-1$ deep neural network (DNN)-based policies of non-leader UAV-BSs.
The leader UAV-BS is responsible for handling communication among multiple non-leader UAV-BSs and collects and distributes information for cooperation among multiple UAV-BSs. The non-leader UAV-BSs transmit state information to the leader UAV-BS, which averages the received information and uses it as input for the next layer.}

Finally, the detailed training procedure can be described in \tblue{Algorithm~\ref{alg:CommNet}}, as follows.
\begin{enumerate}
    \item The weight parameters $\boldsymbol{\theta}^\mu_m$ for \textit{actor} network and $\boldsymbol{\theta}^Q$ for \textit{critic} network are initialized, $\forall m \in [1,M]$ \tblue{(line 1. in Algorithm~\ref{alg:CommNet})}. Note that $\boldsymbol{\theta}^\mu_M$ is parameters of CommNet, and $\{\boldsymbol{\theta}^\mu_m\}^{M-1}_{m=1}$ is originated to DNN.
    \item The both target \textit{actor} and \textit{critic} networks $ \boldsymbol{\theta}^{\hat \mu}_m$ and $\boldsymbol{\theta}^{\hat  Q}$ are initialized (line 2). By using the target network~\cite{mnih2013playing}, the neural network's learning process becomes independent of the parameters. This makes the learning process more stable than the case where the agent UAV-BSs do not use the target network.
    \item All agent UAV-BSs and leader UAV-BS repeat the following procedures until all episodes are over to learn autonomous UAV-BS coordination policies:
    \textit{i)} For every episode, all agent UAV-BSs are doing actions by \textit{actor} networks $\pi_m(a_m|o_m;\boldsymbol{\theta}^\mu_m)$, $\forall m \in [1,M)$. In the leader UAV-BS, actors receives state, \textit{i.e.,} $\pi_M(a_M|s;\boldsymbol{\theta}^\mu_M)$. 
    At each time step, agent UAV-BSs are making transition pairs $\xi = ({S, A, R_{total}, S^{'}})$, where state is recast by $S := \{o_m\}^M_{m=1} \cup s$. 
    After that, they are stored in experience replay buffer $\Phi$. The elements of a transition pair mean the set of states, actions, the reward of agent UAV-BSs, and the observed set of next state spaces, respectively (lines 4–8).
    \textit{ii)} After $\Phi$ is full enough to sample, agent UAV-BSs randomly sample a mini-batch from $\Phi$. 
    That is, the sample size depends on the size of the mini-batch. 
    Note that the training of the neural network proceeds after the buffer is full enough to train to prevent the learning from being focused on the initial experience. 
    This strategy can avoid biased training of the agent UAV-BS by reusing too much initial data.
    \textit{iii)} The $i$-th transition pair of the mini-batch, mean squared Bellman error (\textit{i.e.,} loss function) is calculated with target value $y_{i}$ and $Q(s_{i}, a_{i};\boldsymbol{\theta}^{Q})$ using \textit{Bellman optimality equation} (line 9). The function $Q(\cdot)$ denotes  action-value functions.
    After that, update the \textit{critic} network via descent optimization to reduce value of loss function (line 10-11).
    In target value, $\gamma$ is a discounted factor less than 1, which helps to reach optimal policy faster. The parameters of the \textit{actor} network $\boldsymbol{\theta}^{\mu}_m$, $\forall m \in [1,M]$ are updated via gradient ascent optimization to increase value of objective function (line 12-13).
    \textit{iv)} After updating the parameters of \textit{actor} and \textit{critic} networks, the target parameters $\boldsymbol{\theta}^{\hat{\mu}_m}$ and $\boldsymbol{\theta}^{\hat{Q}}$ are updated at specific periods (line 14).
    \item The training parameters in the neural network (\textit{actor} and \textit{critic}) are shared for agent UAV-BSs. Through this iterative training procedure, agent UAV-BSs have the same parameters for the policy.
\end{enumerate}

\section{Performance Evaluation}\label{sec:ex}

\subsection{Evaluation Setup}
This paper considers the model of UAV-BS as a DJI Phantom4 Pro v2.0 UAV (DJI, Shenzhen, China)~\cite{tvt202106jung}. 
This paper also assumes all UAV-BSs are equipped with 60\,GHz mmWave radio and transceivers~\cite{kim2017feasibility}. UEs are located in a 3D $6,000\,\text{m} \times 6,000\,\text{m} \times 2,500\,\text{m}$ grid map depending on exponential distribution and moving at a maximum speed of $3\,\text{km/h}$ at every time step. 
Each non-agent UAV-BS serves wireless communication service to UEs at the fixed position and malfunctions at $3\%$ probability for each time step. 
Initially, agent UAV-BSs and non-agent UAV-BSs are located at the grid's center, and $3,000\,\text{m}$ apart from the center, respectively. Each episode has a total of 40-time steps (=$30\,$minute) where each time step is 45 seconds. 

In this paper, there are two neural network structures, CommNet and DNN, which consist of six dense layers. The number of nodes in the first to the last layer is 64, and we use the ReLU function in the first to fifth layers. We use a Xavier initializer for initializing weight and an Adam optimizer for minimizing loss. In addition, we use the $\epsilon$-greedy method to make the agent UAV-BSs explore to experience a variety of actions. Additional environment setup parameters are summarized in Table \ref{tab:setup}.

To investigate whether our proposed method achieves a common goal of agent UAV-BSs, we benchmark our proposed method with three comparisons as follows.
\begin{itemize}
    \item \textit{Proposed Method:} Among $M$ agent UAV-BSs, one agent UAV-BS is a CommNet-based leader UAV-BS. The remaining $M-1$ agent UAV-BSs are DNN-based agents, providing the leader agent UAV-BS with their own rich experiences. 
    The leader UAV-BS provides its communication variables to other agent UAV-BSs by averaging other agent UAV-BSs' hidden variables, \textit{i.e.,} experiences.
    \item \textit{Random Method:} All agent UAV-BSs have random policies that randomly choose their actions, \textit{i.e.,} random walk. 
    Since agent UAV-BSs have random policies, they do not require to observe their state information or communication between agent UAV-BSs.
    \item \textit{Comp1 Method:} There is no leader UAV-BS. In other words, all agent UAV-BSs try to provide optimal mmWave access to UEs based on DNN. Since no CommNet-based agent UAV-BSs exist, there are no inter-agent UAV-BS communications. Thus, each agent UAV-BS makes an action decision as if it were in a single-agent UAV-BS environment.
    For training Comp1, independent Q-learning is used \cite{tan1993multi}. 
    \item \textit{Comp2 Method:} All agent UAV-BSs try to provide optimal mmWave access to UEs based on CommNet. All agent UAV-BSs share their experiences at the same time. The redundancy between communication variables is high, \textit{i.e.,}  unnecessary computation costs may occur.
\end{itemize}

\begin{table}[t!]
\centering
\caption{Environmental/Experimental Setup Parameters}
\renewcommand{\arraystretch}{1.0}
\begin{tabular}{l||r}
\toprule[1pt]
\textbf{\textsf{Parameters}} & \textbf{\textsf{Value}} \\ \midrule
Time steps per episode, $T$ & 40 \\
The number of agent UAV-BSs, $M$ & 4 \\
The number of UEs, $N$ & 25 \\
The number of non-agent UAV-BSs, $K$ & 3 \\
The Number of dense layer, $L$ & 6 \\
Altitude of agent UAV-BSs, $z$ & [1500, 2000, 2500] m \\
UAV's directional antenna beamwidth, $\theta_W$ & $80^\circ$\\
Optimizer & Adam \\
Learning rate & 0.001 \\
Target network update cycle & 20 \\
Training epochs & 50k \\
\bottomrule[1pt]
\end{tabular}
\label{tab:setup}
\end{table}

\subsection{Evaluation Results}
This section evaluates the performance of the above four methods. We benchmark our proposed method with various metrics, \textit{i.e.,} \textit{i)} reward convergence, \textit{ii)} user connectivity, \textit{iii)} service reliability, \textit{iv)} computational cost, and \textit{v)} trained behavior of agent UAV-BSs per episode.

\subsubsection{Reward Convergence in POMDP}
\begin{figure}[t!]
    \centering
    \ \ \ \ \ \includegraphics[width=0.6\linewidth]{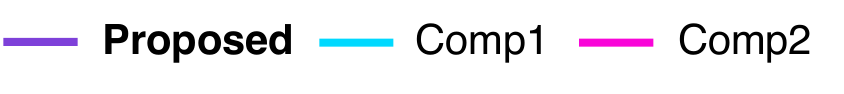}
    \subfigure[POMDP Environment.]{
    \includegraphics[width=0.45\linewidth]{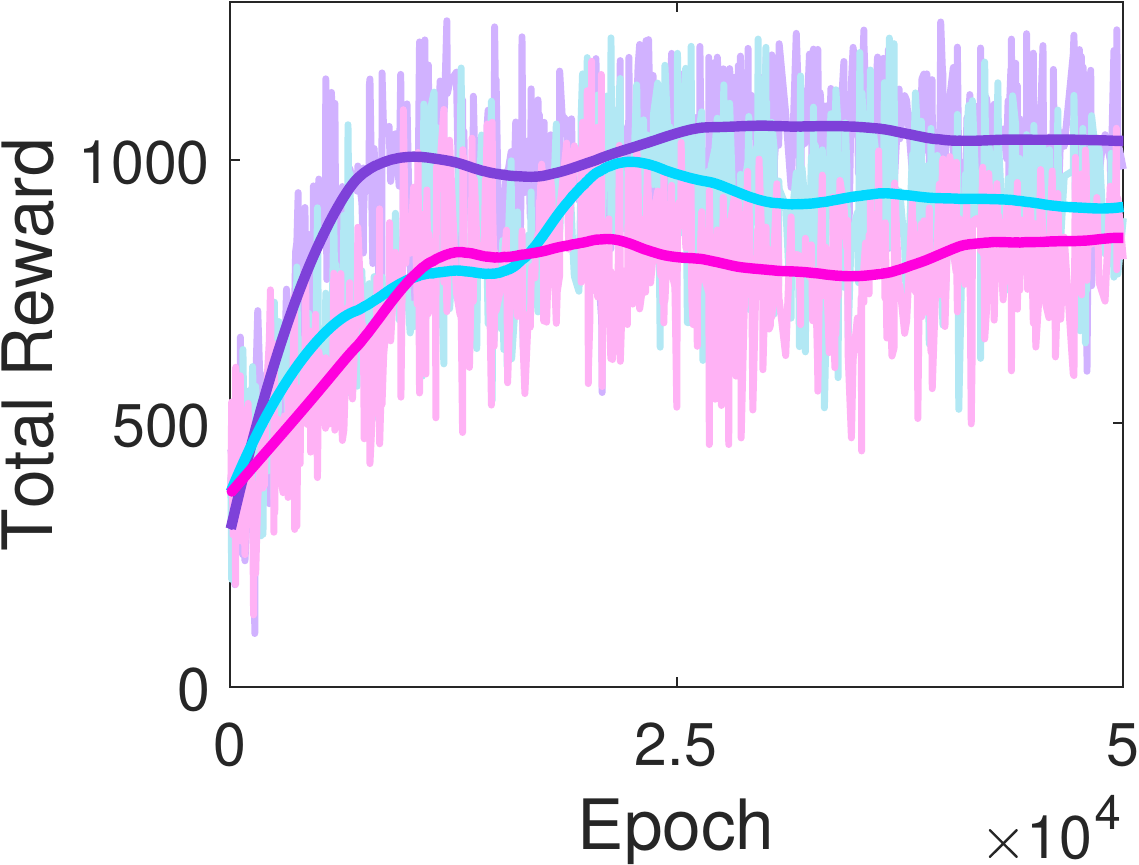}
    \label{fig:POMDP}
    }
    \subfigure[FOMDP Environment.]{
    \includegraphics[width=0.45\linewidth]{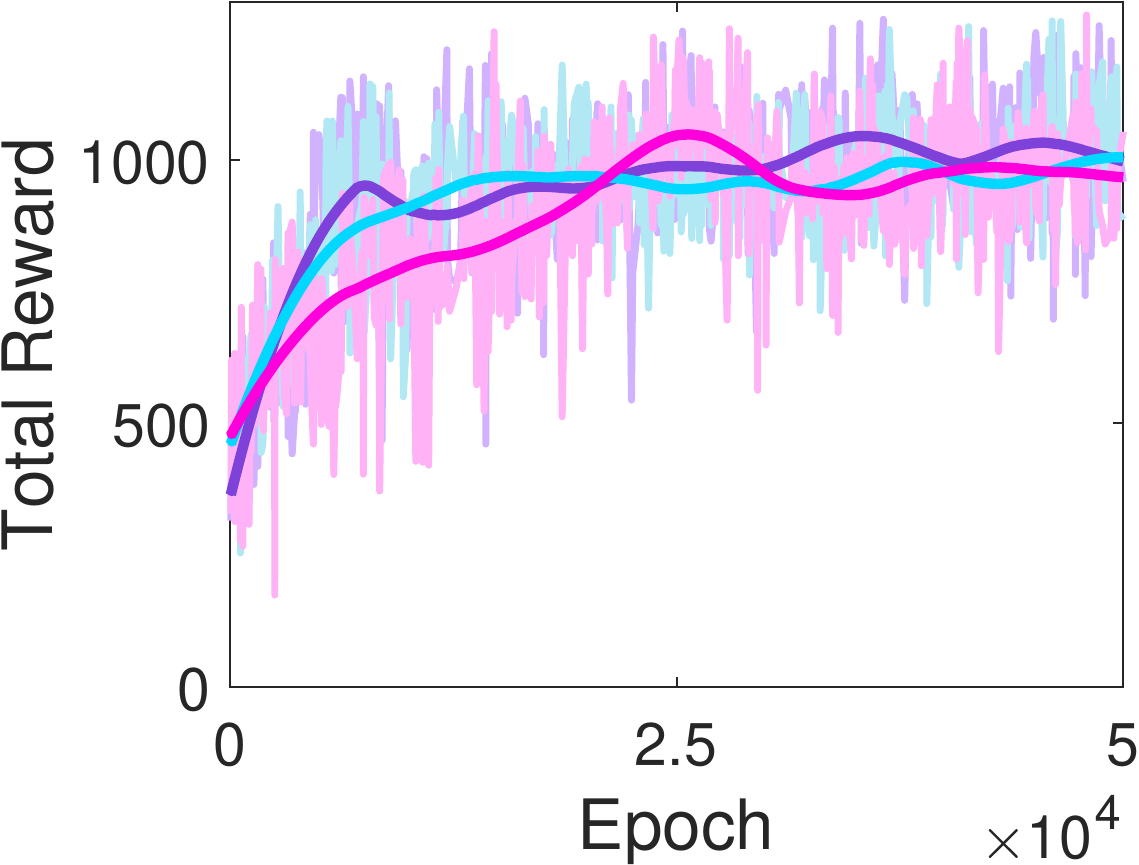}
    \label{fig:FOMDP}
    }\\
    \caption{Comparison total reward of each method with POMDP and FOMDP. Note that the FOMDP environment cannot be reflective of reality when considering the actual specifications of UAVs.}
    \label{fig:POMDP and FOMDP}
\end{figure}

In a natural multi-agent UAV-BS environment, each agent UAV-BS observes different information and has a limited scope of observation. We consider both POMDP and FOMDP environments for the generalizability.
In this case, CommNet helps multiple agent UAV-BSs to jointly achieve a common goal in a partially observable Markov decision process (POMDP). In a POMDP, each agent UAV-BS can only observe a specific range of environmental information based on its location.
On the other hand, in a fully observable Markov decision process (FOMDP), communication between agent UAV-BSs is not required. This is because every agent UAV-BS knows all environment information (\textit{e.g.,} location of other agent UAV-BSs and UEs, coverage of other agent UAV-BSs).
Fig.~\ref{fig:POMDP and FOMDP} presents the total reward for the proposed, Comp1 and Comp2 methods in each situation of POMDP and FOMDP. Fig.~\ref{fig:POMDP} connotes three significant results, \textit{i.e.,} reward value, convergence speed, and stability.

\begin{itemize}
    \item \textit{Reward Value:} The proposed method has the highest reward at about 1,000 in all epochs. The reward of Comp1 follows the next at around 900, and Comp2 converges to the smallest reward value at about 800. Therefore, the proposed method has achieved the near-optimal compared to other comparison methods.
    Note that the proposed method's DNN-based agent UAV-BSs provide abundant experiences to the leader UAV-BS, a CommNet-based agent UAV-BS. 
    These help multiple agent UAV-BSs to get optimal policies with high rewards.

    \item \textit{Convergence Speed and Stability:} 
    In the proposed method, the reward value converges to about 1,000 from about 8,500 epochs to the end of learning, which corresponds to the fastest reward convergence. 
    In the case of the Comp1 method, the reward increases to approximately 1,000 at about 22,000 epochs. 
    For the Comp2 method, the reward value converges to about 800 from around 12,500 epochs to the end. 
    The proposed method and Comp2 method, which have at least one CommNet-based agent UAV-BS, show relatively higher convergence speed and stability than Comp1. CommNet allows multiple agent UAV-BSs to have correlated policies with each other. In other words, all agent UAV-BSs cooperatively try to have optimal policies, reducing the reward fluctuations in learning. As a result, the CommNet-based agent UAV-BS helps itself and other non-leader agent UAV-BSs to learn policies more reliably.
\end{itemize}

As shown in Fig.~\ref{fig:FOMDP}, the total rewards of all methods converges from 900 to 950 in all methods. 
In other words, communication using CommNet is meaningless in a FOMDP environment. This is because the agent UAV-BS already knows all the environmental information of the FOMDP environment.
The Comp1 method can have a total reward value similar to the proposed method and the Comp2 method without a CommNet-based agent UAV-BS.

\begin{figure}[!t]
    \centering
    \includegraphics[width=1\linewidth]{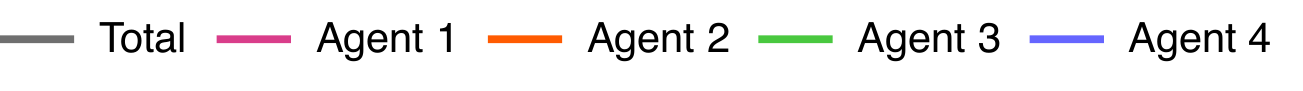}\\
    \subfigure[\textbf{Proposed.}]{
    \includegraphics[width=0.45\linewidth]{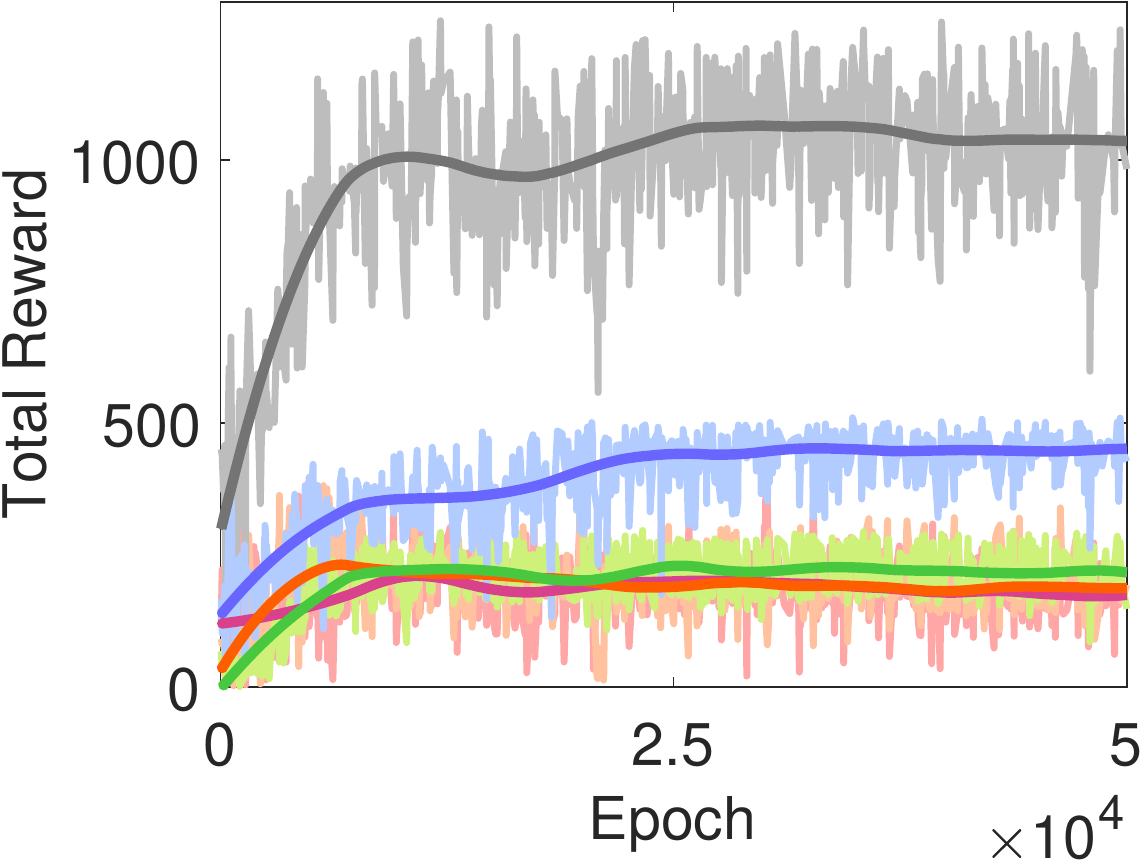}
    \label{fig:Proposed reward}
    }
    \subfigure[Random.]{
    \includegraphics[width=0.45\linewidth]{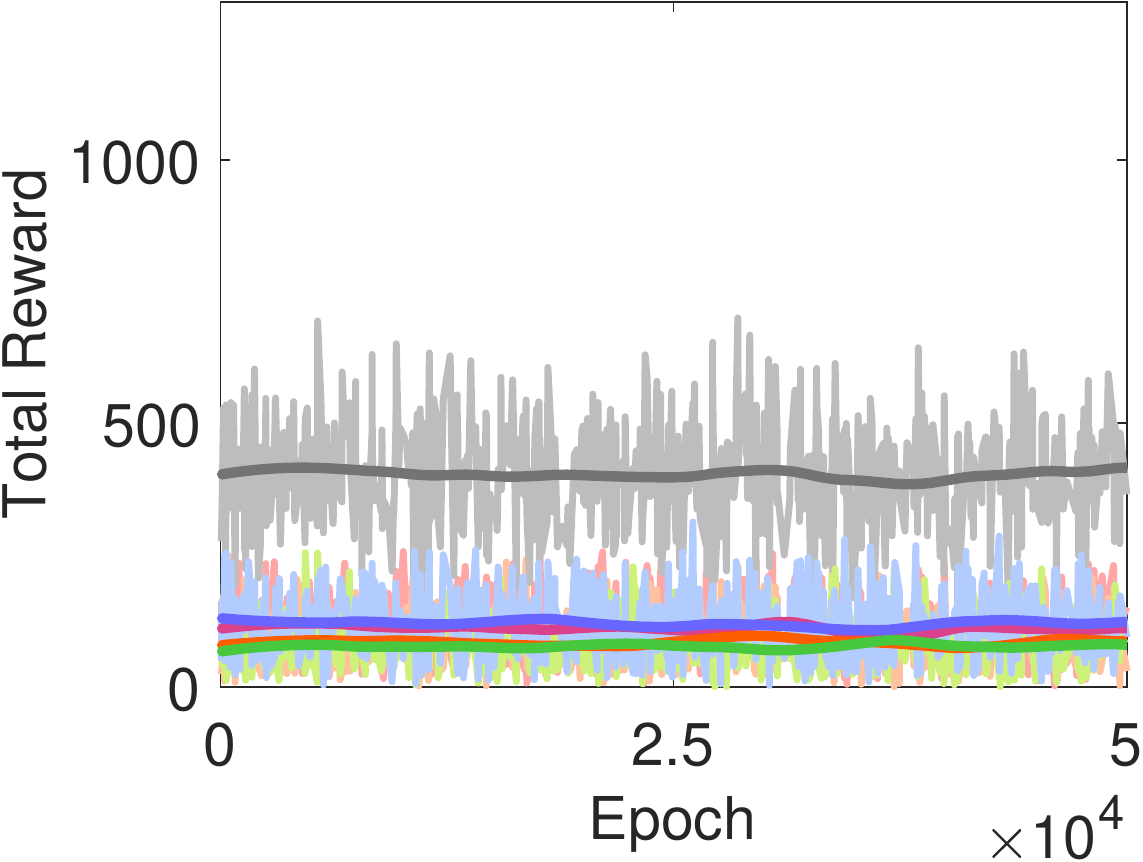}
    \label{fig:Random reward}
    }\\
    \subfigure[Comp1.]{
    \includegraphics[width=0.45\linewidth]{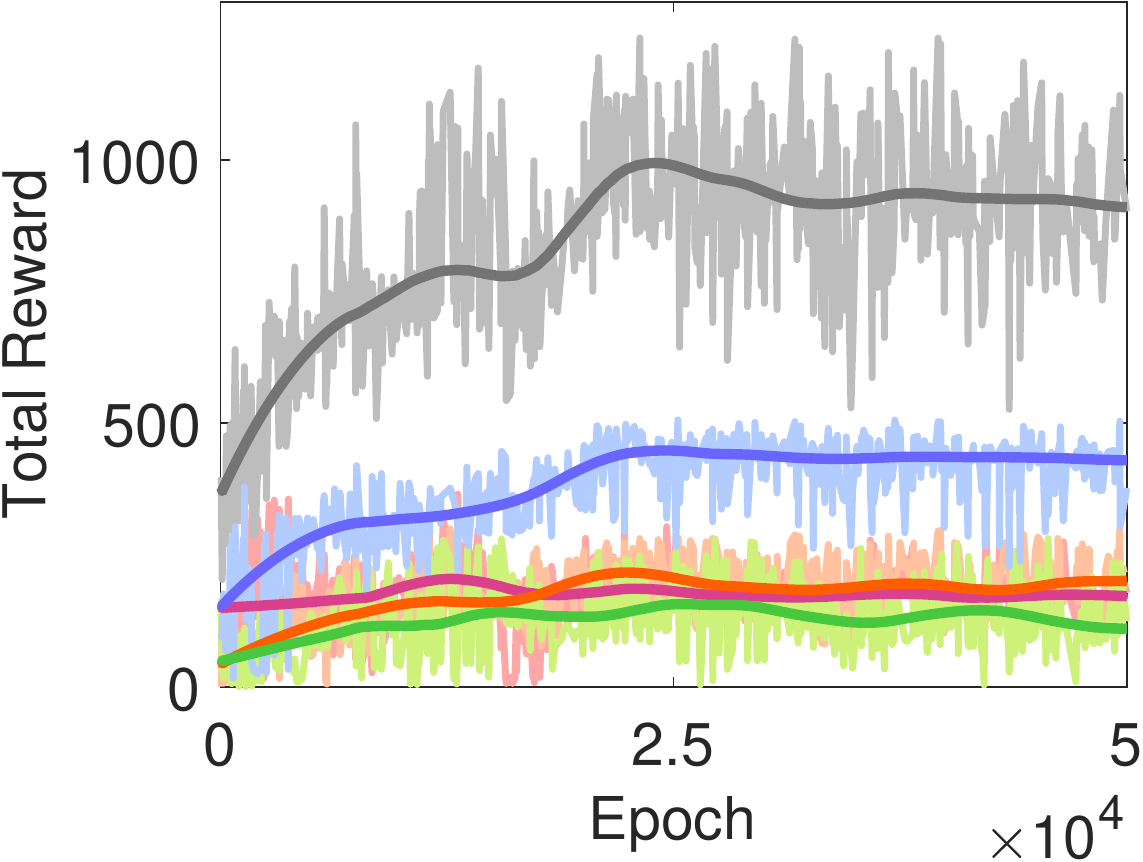}
    \label{fig:Comp1 reward}
    }
    \subfigure[Comp2.]{
    \includegraphics[width=0.45\linewidth]{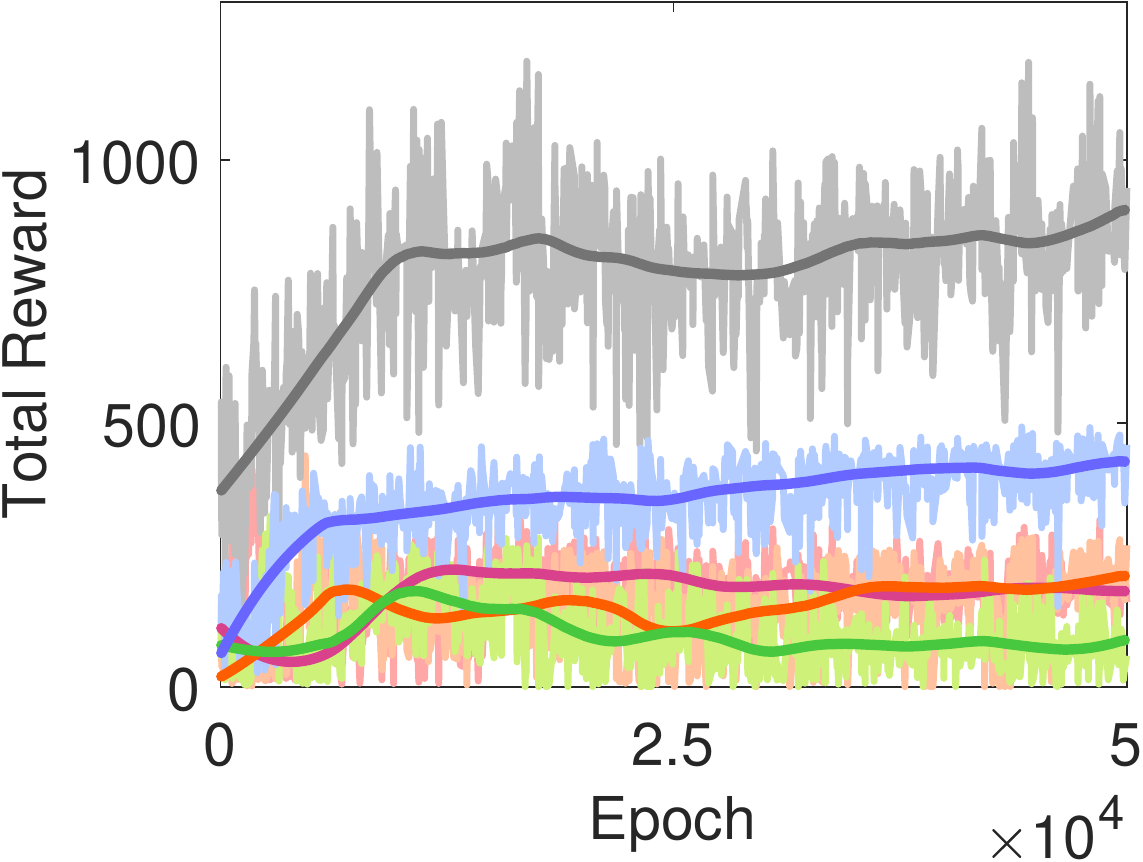}
    \label{fig:Comp2 reward}
    }
    \caption{Total and each agent UAV-BS's reward in all methods in the POMDP.}
    \label{fig:each agent's reward}
\end{figure}

Fig.~\ref{fig:each agent's reward} presents the performance of each agent UAV-BS in all methods in a POMDP environment. Except for Fig.~\ref{fig:Random reward}, the reward value of $b_4$ is commonly the highest, from about 400 to 450 among four agent UAV-BSs. In Fig.~\ref{fig:Proposed reward}, the remaining three DNN-based agent UAV-BSs, which are non-leader agent UAV-BSs, get even reward at around 200. In the Comp1 and Comp2 method, the reward of $b_1$ and $b_2$ follows the next at around 200. However, $b_3$ has a lower reward than $b_1$ and $b_2$ in common. In addition, two methods finally get the second highest total reward value to about 900. While the proposed method shows the fastest convergence speed and stability, the Comp1 shows the more unstable convergence due to the reward fluctuation, and the Comp2 shows the slower convergence speed compared to the proposed method, respectively. Consequently, we validate that the proposed method shows the highest reward value and outstanding reward convergence performance among all methods in the POMDP environment.

\begin{figure}[!ht]
    \centering
    \ \ \ \ \includegraphics[width=0.8\linewidth]{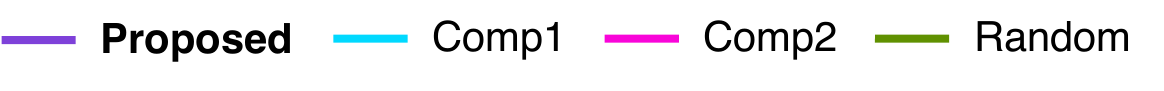}
    \subfigure[Total Support Rate.]{
    \includegraphics[width=0.45\linewidth]{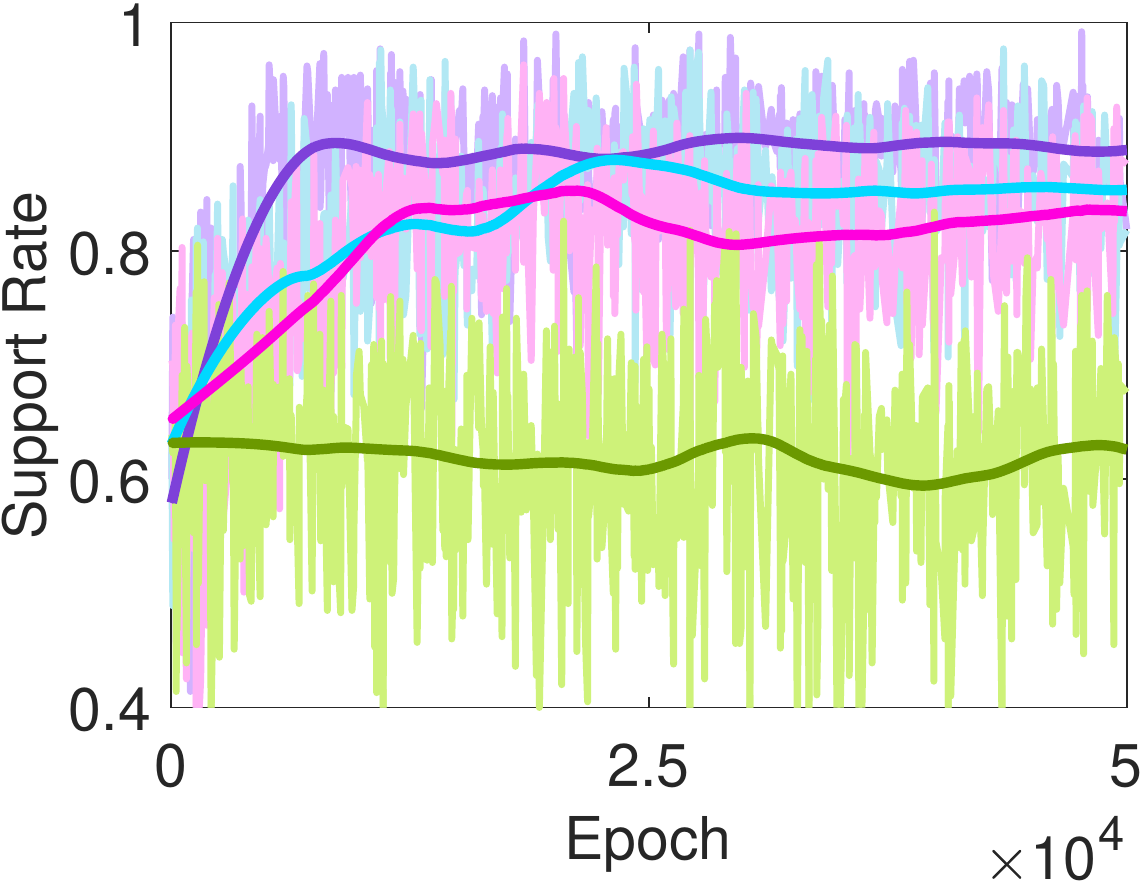}
    \label{fig:Support rate}
    }
    \subfigure[Total QoS.]{
    \includegraphics[width=0.45\linewidth]{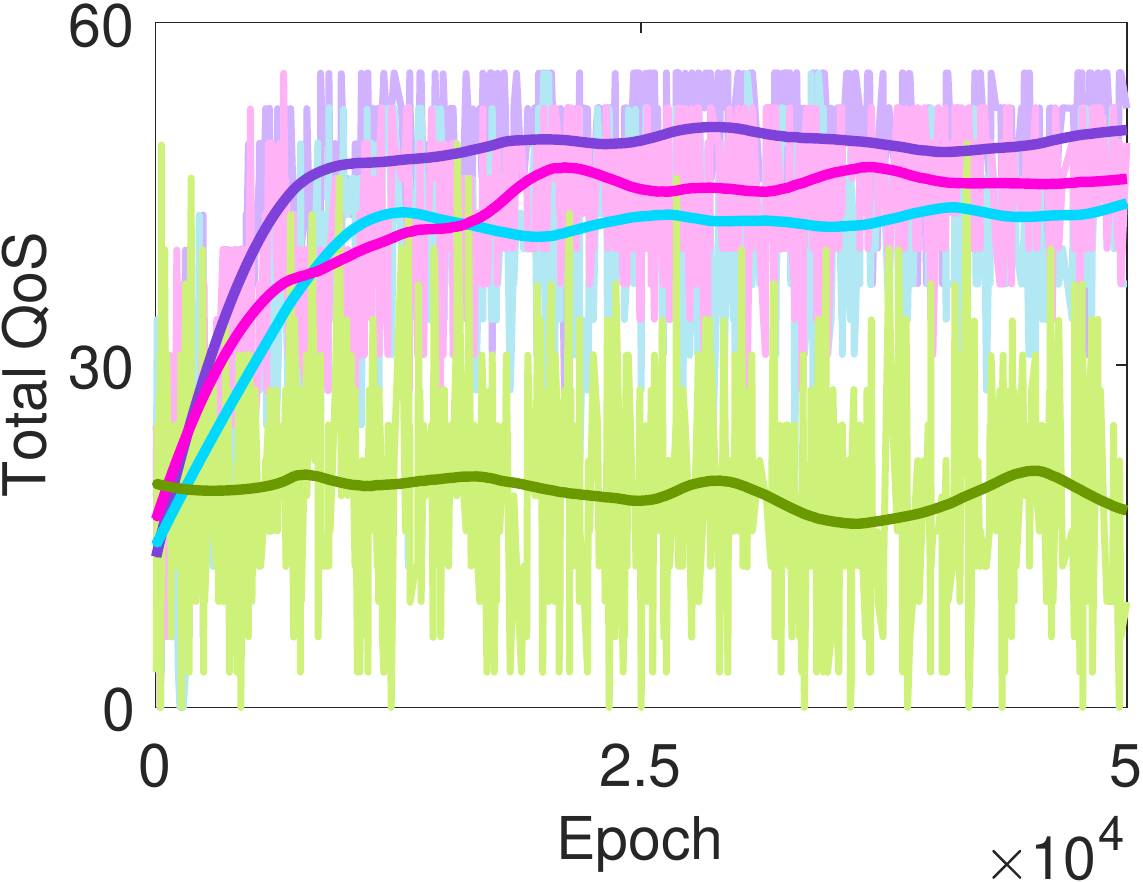}
    \label{fig:total Quality}
    }
    \caption{Total support rates and total QoS during the training phase}.
\end{figure}

\subsubsection{User Connectivity}
\begin{figure}[!t]
    \centering
    \ \ \ \ \includegraphics[width=0.75\linewidth]{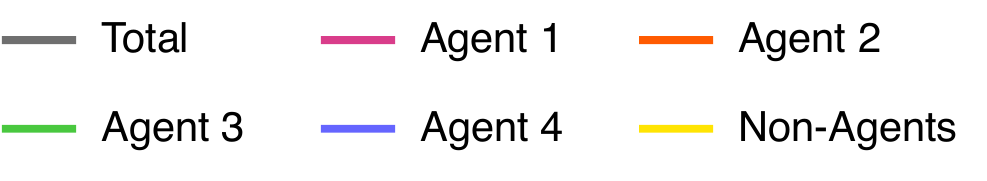}\\
    \subfigure[\textbf{Proposed.}]{
    \includegraphics[width=0.45\linewidth]{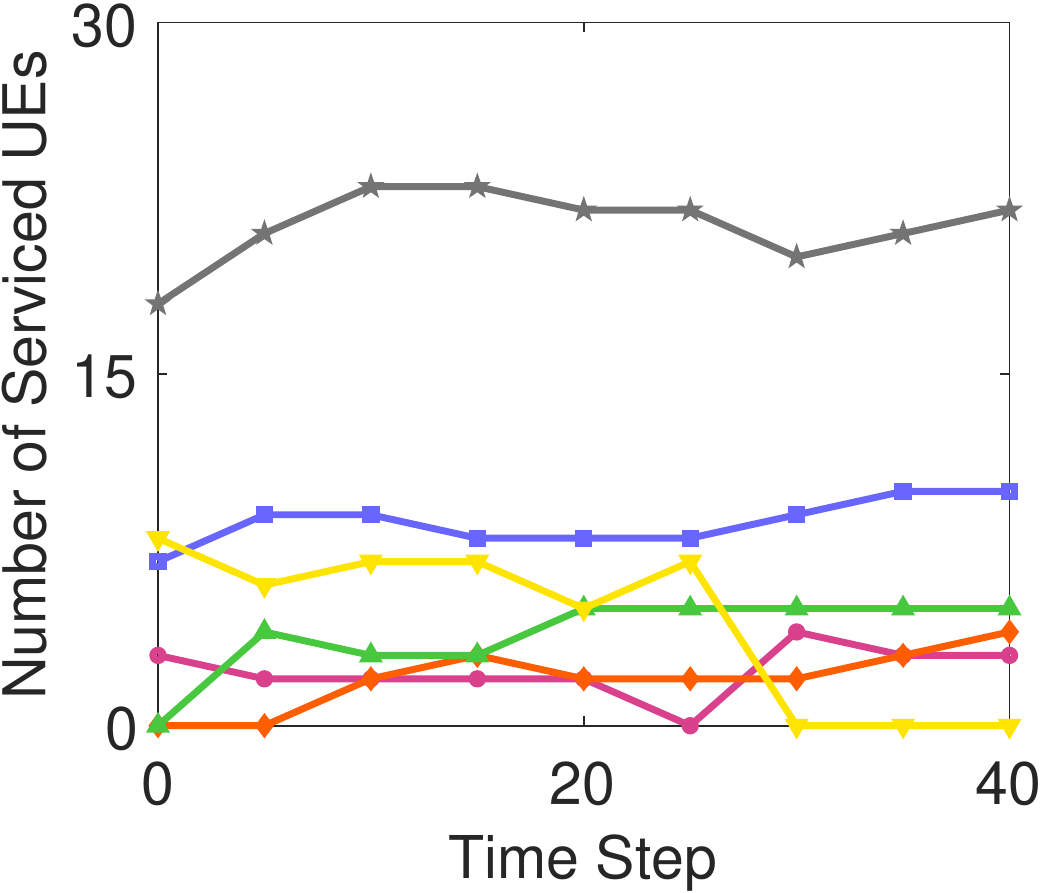}
    \label{fig:numUEs in proposed}
    }
    \subfigure[Random.]{
    \includegraphics[width=0.45\linewidth]{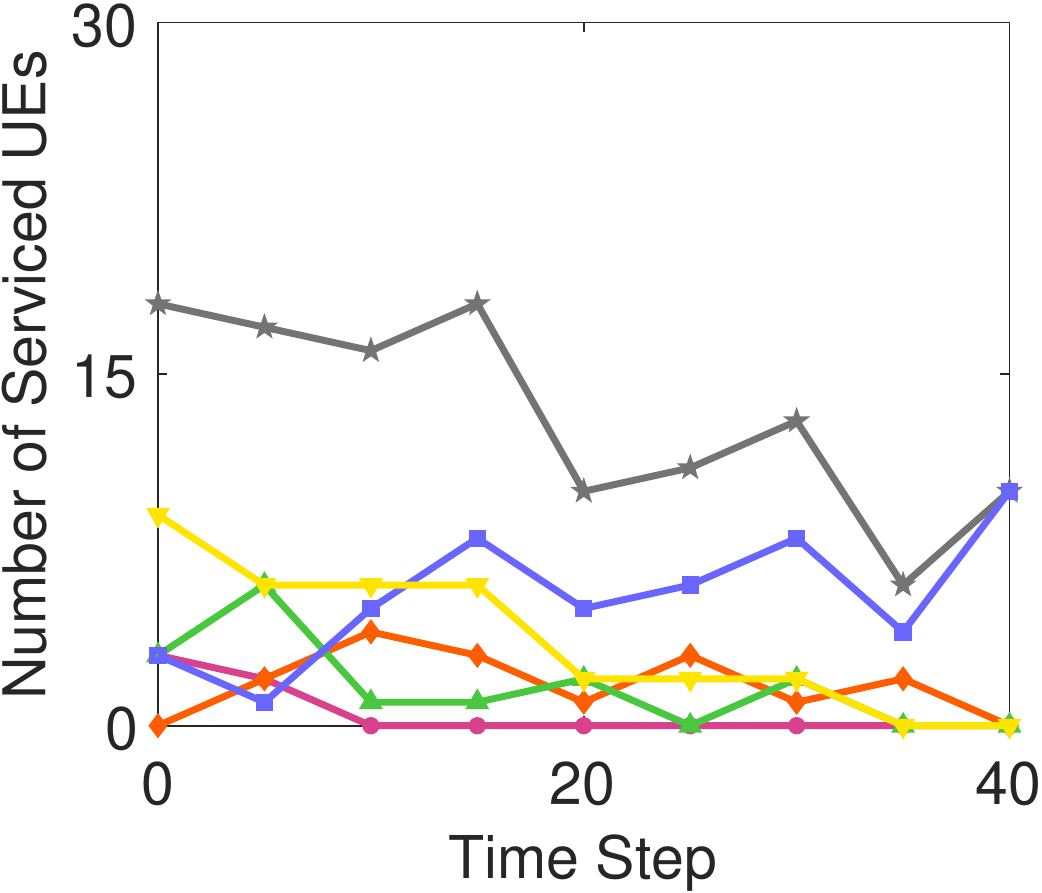}
    \label{fig:numUEs in Random}
    }\\
    \subfigure[Comp1.]{
    \includegraphics[width=0.45\linewidth]{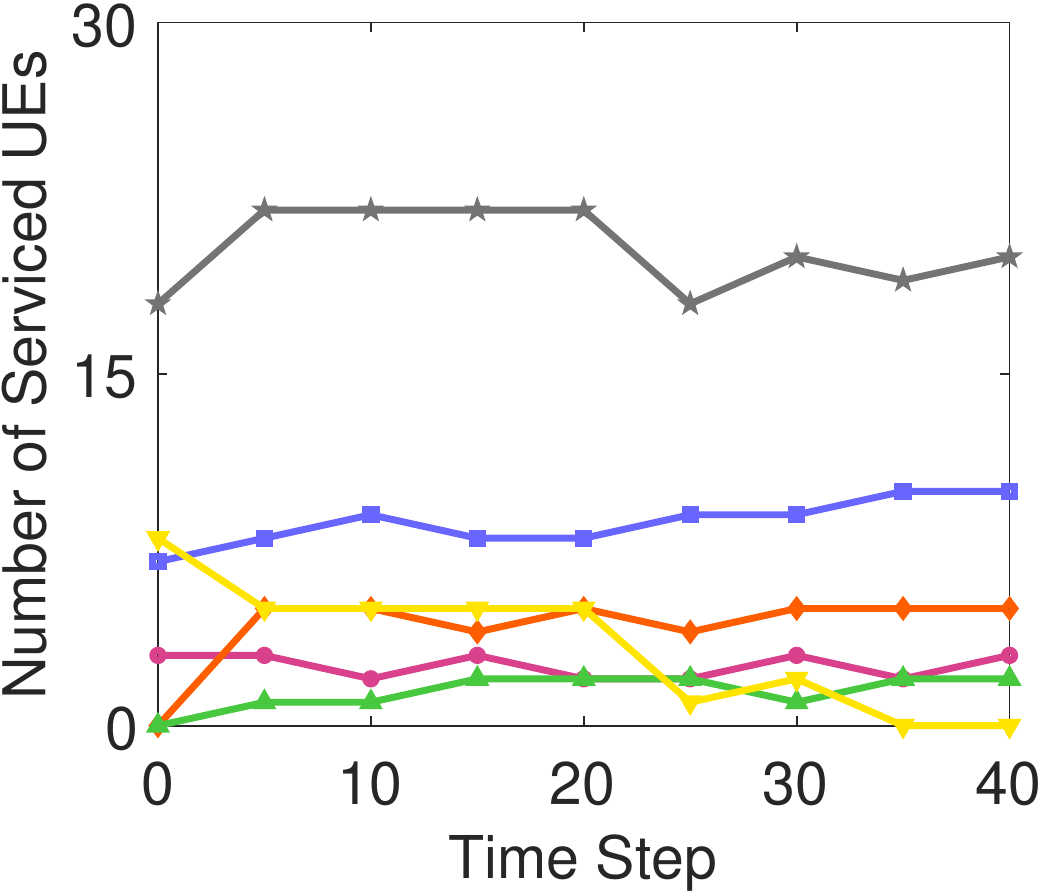}
    \label{fig:numUEs in Comp1}
    }
    \subfigure[Comp2.]{
    \includegraphics[width=0.45\linewidth]{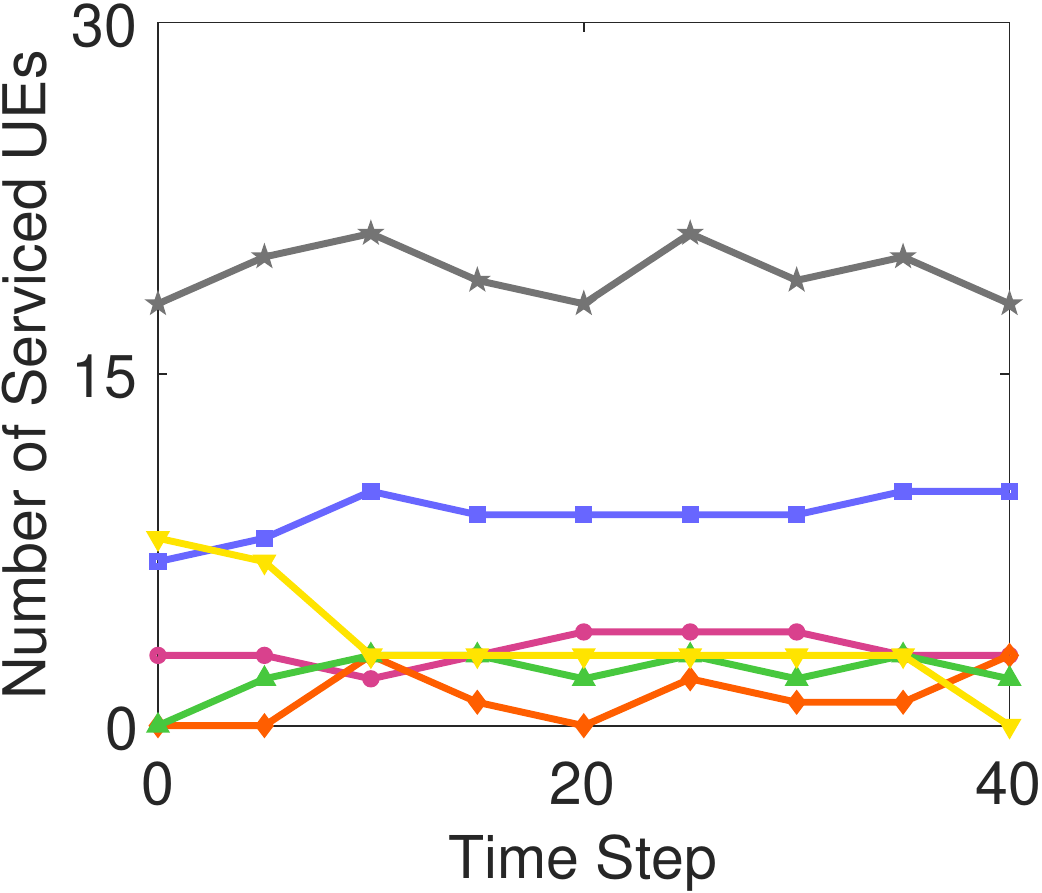}
    \label{fig:numUEs in Comp2}
    }\\
    \caption{Number of UEs serviced by UAV-BSs during the inference phase.}
    \label{fig:numUEs}
\end{figure}

This section evaluates user connectivity in terms of support rate in Fig.~\ref{fig:Support rate}, the number of UEs receiving mmWave communications in Fig.~\ref{fig:numUEs} and Table~\ref{tab:numUEs}.

\BfPara{Training Phase}
The support rate of the proposed method is the highest in almost all epochs, and its value is about 0.9. 
In addition, the increasing rate of the proposed method is fastest among comparisons. The support rate increases to about 0.9 at around 7,500 epochs. 
After that, the support rate is consistently about 0.9 until the end of the learning. In the Comp1 and Comp2 methods, the support rate finally converges to about 0.8. 
Significantly, the support rate increases to about 0.9 at around 2,300 epochs in the Comp1 method. However, it decreases to 0.8 due to the reward fluctuation of Comp1. 
The random method shows the worst support rate of all methods, where the value is about 0.6. As a result, the proposed method presents the highest value, the fastest increasing rate, and stability compared to the other methods.

\begin{table*}[ht]
    \centering
    {\footnotesize
\caption{The number of UEs serviced by UAV-BSs at each time step as the episode progresses during the inference phase.}
\renewcommand{\arraystretch}{1.0}
\begin{tabular}{c||ccc|ccc|ccc} \toprule[1pt]
\textbf{\textsf{Time}} & \multicolumn{3}{c}{\textbf{\textsf{Proposed}}} & \multicolumn{3}{|c|}{\textbf{\textsf{Comp1}}} & \multicolumn{3}{c}{\textbf{\textsf{Comp2}}} \\
   \textbf{\textsf{Step}} & Total & Agents & Non-Agent & Total & Agents & Non-Agents & Total & Agents & Non-Agent \\ \midrule
$t = 0$ & 18 & 10 & 8 & 18 & 10 & 8 & 18 & 10 & 8 \\ 
$t = 5$ & 21 & 15 & 6 & 22 & 17 & 5 & 20 & 13 & 7 \\ 
$t = 10$ & 23 & 16 & 7 & 22 & 17 & 5 & 21 & 18 & 3 \\ 
$t = 15$ & 23 & 16 & 7 & 22 & 17 & 5 & 19 & 16 & 3 \\ 
$t = 20$ & 22 & 17 & 5 & 22 & 17 & 5 & 18 & 15 & 3 \\ 
$t = 25$ & 22 & 15 & 7 & 18 & 17 & 1 & 21 & 18 & 3 \\ 
$t = 30$ & 20 & 20 & 0 & 20 & 18 & 2 & 19 & 16 & 3 \\ 
$t = 35$ & 21 & 21 & 0 & 19 & 19 & 0 & 20 & 17 & 3 \\ 
$t = 40$ & 22 & 22 & 0 & 20 & 20 & 0 & 18 & 18 & 0 \\ \midrule
Total & \textbf{192} & \textbf{152} & \textbf{40} & 183 & 152 & 31 & 174 & 141 & 33 \\
\bottomrule[1pt]
\end{tabular}
\label{tab:numUEs}
}
\end{table*}

\BfPara{Inference Phase}
Fig.~\ref{fig:numUEs} shows the results of user connectivity. 
Note that `Remain' means that the number of supported UEs from non-agent UAV-BSs.
As shown in Fig.~\ref{fig:numUEs}(a)--(c), we can see that $b_1$ provides mmWave communication to most UEs.
Fig.~\ref{fig:numUEs in Random} shows that the number of serviced UEs is the smallest. In other words, agent UAV-BSs in the random method show the weakest user connectivity.
As shown in Fig.~\ref{fig:numUEs}, and Table~\ref{tab:numUEs}, all UAVs are malfunctioned at $t\in(25, 30],\;t\in(30, 35],\;$and$\;t\in(35, 40]$ in the proposed method, Comp1, and Comp2, respectively.
Although all non-agent UAVs malfunction at the fastest speed with the proposed scheme, the number of UEs serviced by UAV-BSs is the largest compared to Comp1 and Comp2. Note that the total number of UEs serviced by agent UAV-BSs is equal in the proposed method and Comp1 method despite all non-agent UAVs in proposed scheme malfunction earlier than other schemes. Besides, agent UAV-BSs can cover 20--22 UEs after all UAVs malfunction in the proposed method, whereas, in the Comp1 and Comp2 methods, agent UAV-BSs can only cover 19--20 UEs and 18 UEs, respectively.  
To sum it up, agent UAV-BSs in the proposed method communicate efficiently with each other and have more abundant experience than other methods. Furthermore, agent UAV-BSs in the proposed method show the strongest user connectivity.

\subsubsection{Service Reliability}
This section evaluates wireless communication service reliability through the QoS of UEs and the energy consumption of each agent UAV-BS.

\BfPara{QoS}
%
%
Fig.~\ref{fig:total Quality} represents the total QoS of UEs receiving mmWave service of each method in the training phase. Except for the very early stage of training, UEs in the proposed method gets the largest QoS from UAV-BSs at all learning times. The proposed method shows the fastest increase rate until 9k epochs, the most stable convergence rate at around 50, and minuscule fluctuations.
\begin{figure}[!t]
    \centering
    \includegraphics[width=1\linewidth]{New_Figure/total-agent-description.pdf}\\
    \subfigure[\textbf{Proposed.}]{
    \includegraphics[width=0.45\linewidth]{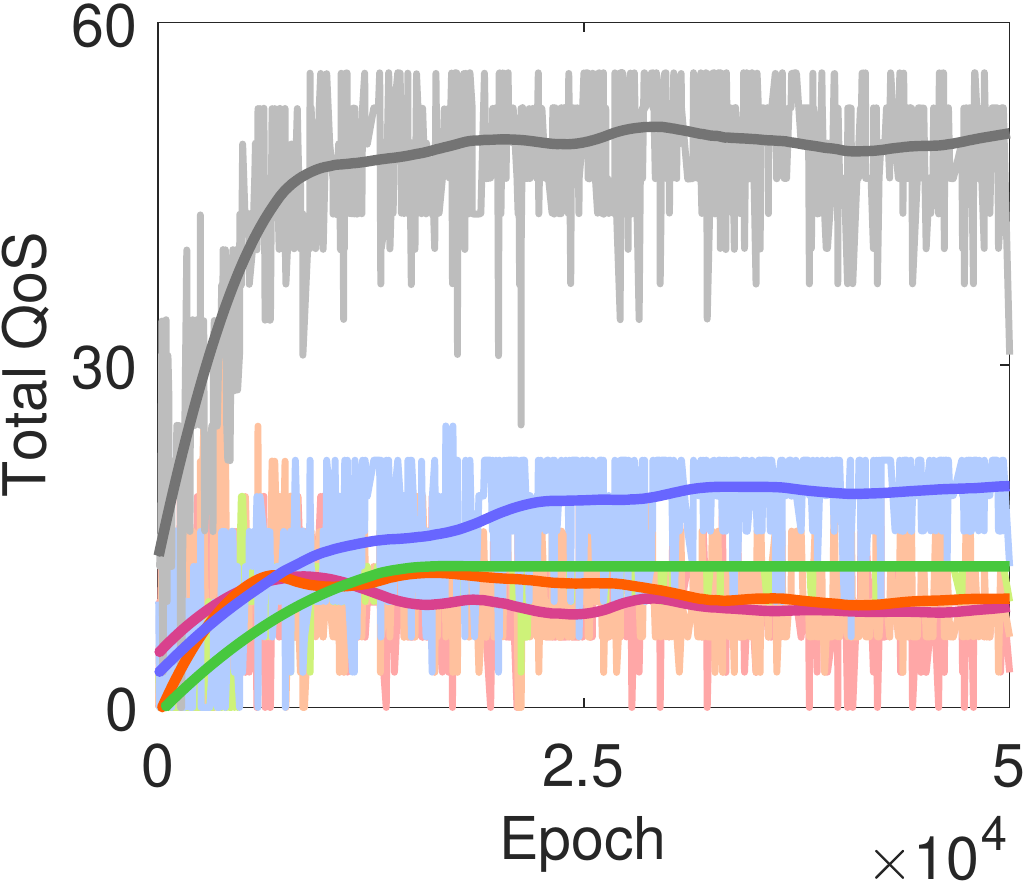}
    }
    \subfigure[Random.]{
    \includegraphics[width=0.45\linewidth]{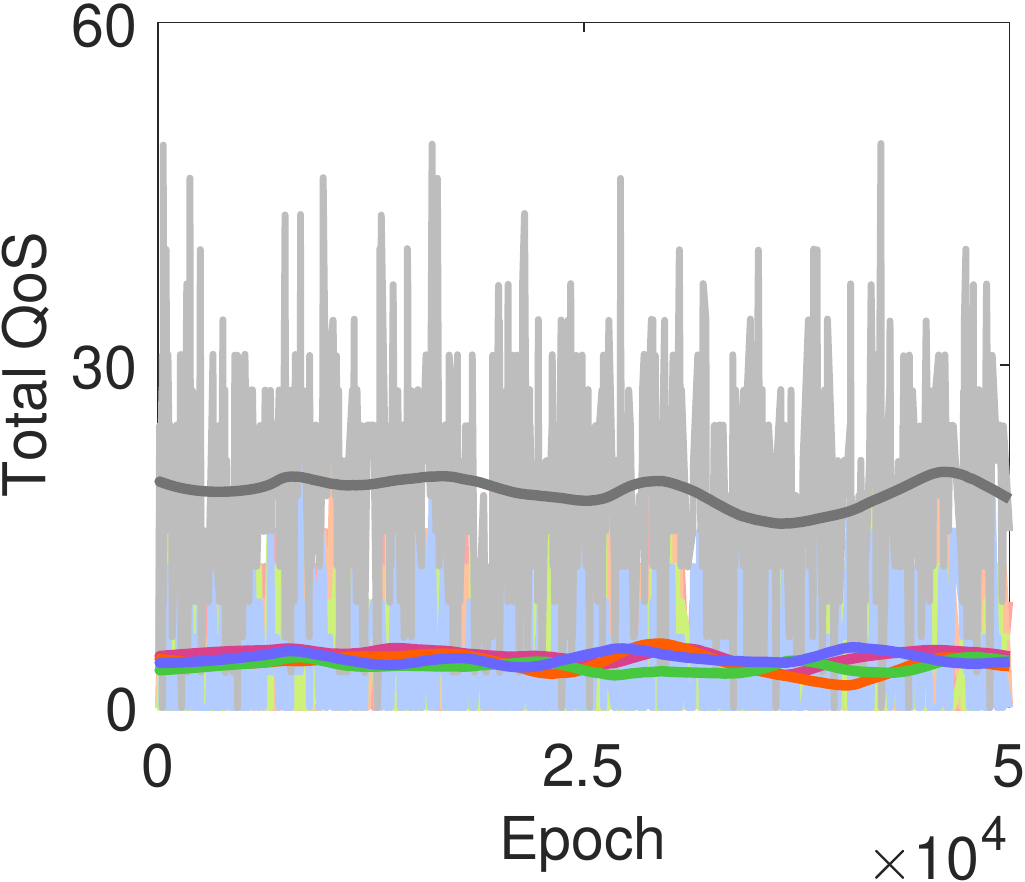}
    }\\
    \subfigure[Comp1.]{
    \includegraphics[width=0.45\linewidth]{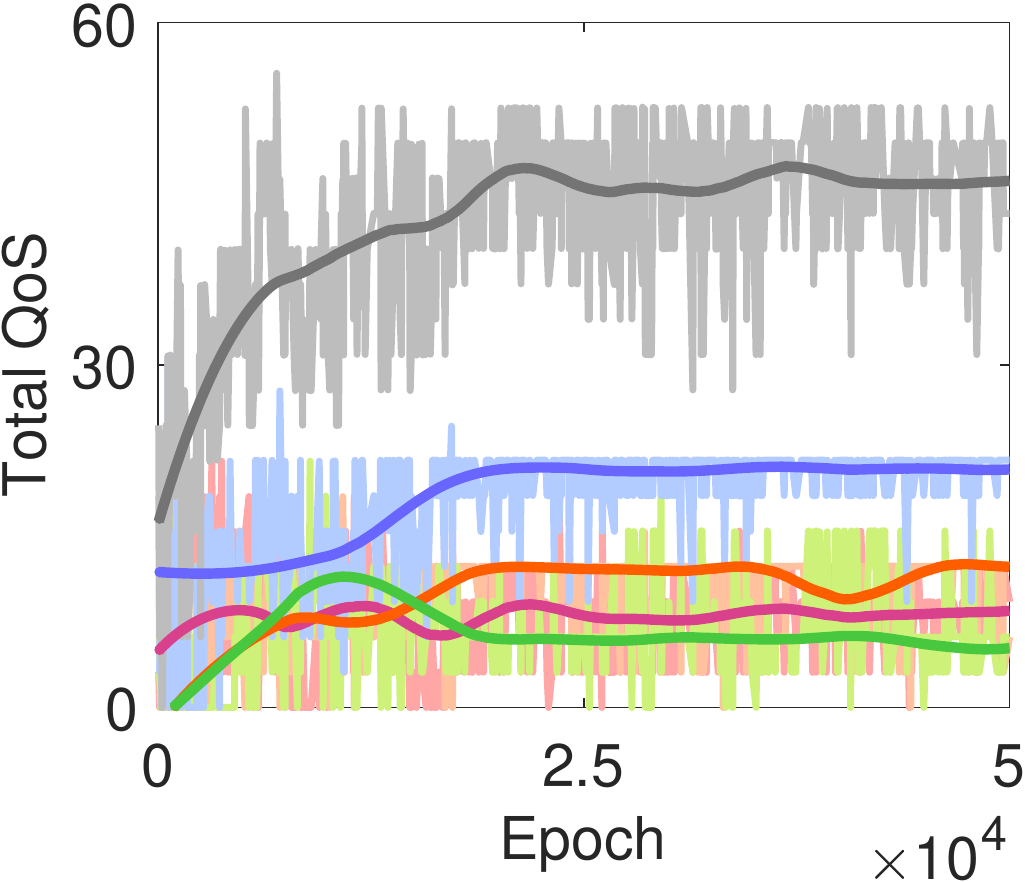}
    }
    \subfigure[Comp2.]{
    \includegraphics[width=0.45\linewidth]{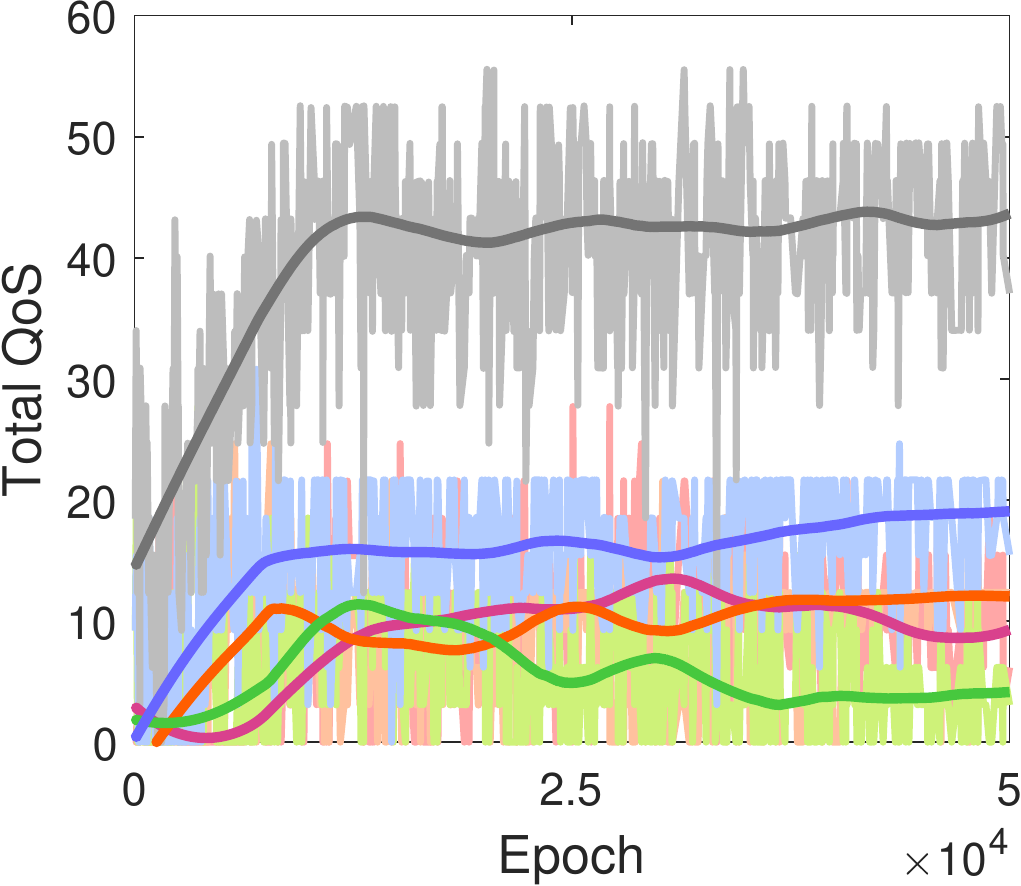}
    }
    \caption{Each agent UAV-BS's serving QoS.}
    \label{fig:Quality}
\end{figure}
For all methods in Fig.~\ref{fig:Quality}, the aspect of QoS values provided by each agent UAV-BS is similar to Fig.~\ref{fig:numUEs}. Based on the various experiences of non-leader agent UAV-BSs and the inter-agent UAV-BS communication of the leader agent UAV-BS, agent UAV-BSs of the proposed method provide relatively evenly QoS to UEs compared to the Comp1 and Comp2 methods. In the case of the random method, even if agent UAV-BSs deliver QoS to UEs evenly, the overall QoS of UEs is very small. Thus, the mmWave service provided by agent UAV-BSs in the random method is not good.
\begin{figure}[ht]
    \centering
    \ \ \ \ \ \includegraphics[width=0.6\linewidth]{legend.pdf}
    \subfigure[Total overlapping rate.]{
    \includegraphics[width=0.4\linewidth]{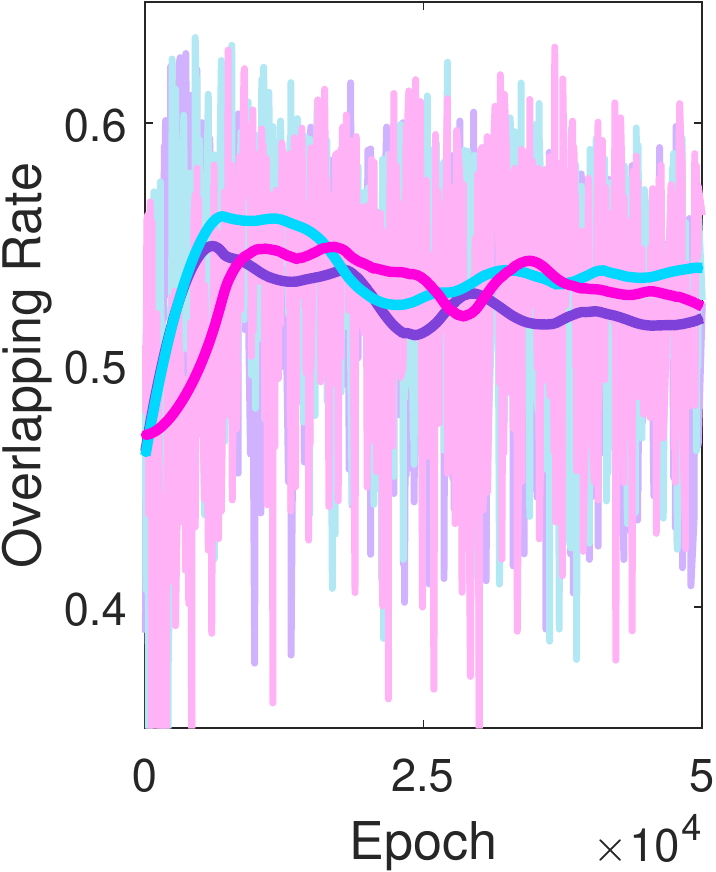}
    \label{fig:Total overlapping rate}
    }
    \subfigure[Average overlapping rate.]{
    \includegraphics[width=0.4\linewidth]{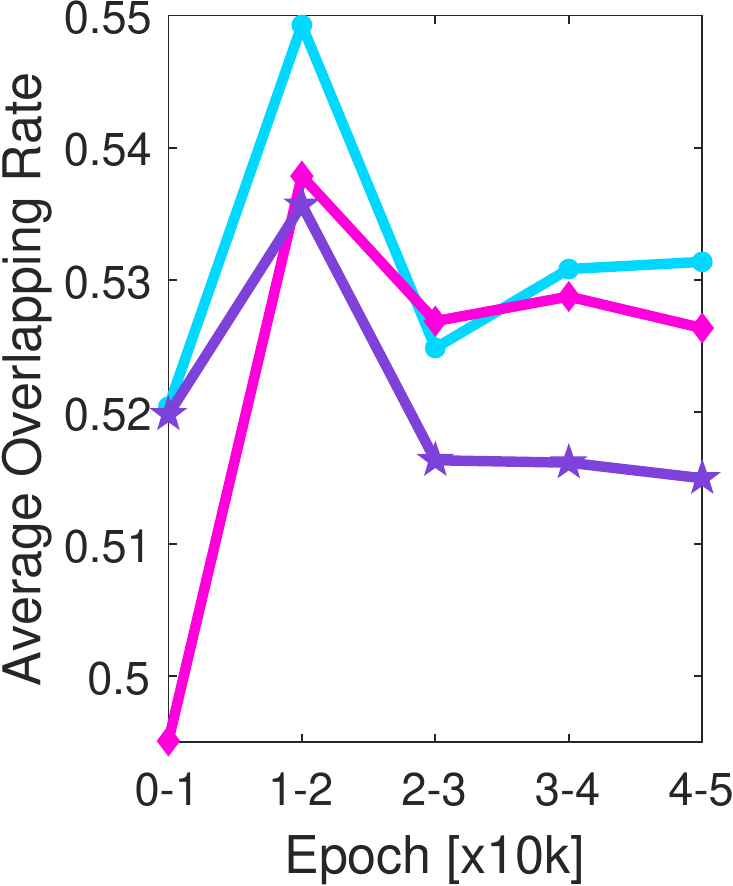}
    \label{fig:Average overlapping rate}
    }
    \caption{Total and average overlapping rates in each method.}
    \label{fig:Overlapping rate}
\end{figure}

Next, we investigate the impact on QoS by reducing overlapping area. 
In order to provide high QoS to many UEs, UAV-BSs should avoid overlapping as much as possible to reduce interference from other UAV-BSs, according to \eqref{eq:interference} and \eqref{eq:Shannon's}.
Fig.~\ref{fig:Overlapping rate} shows the overlapping rate of the entire UAV-BSs during the training phase, as well as the average overlapping rate in units of 10k, also during the training phase. In Fig.~\ref{fig:Total overlapping rate}, the overlapping degree in the proposed method is not the greatest in all epochs. Furthermore, the finally converged overlapping degree is the smallest among all the methods except the random method, where UAV-BSs serve the poorest service. For a more intuitive comparison, this paper also suggests Fig.~\ref{fig:Average overlapping rate}. The overlapping rate of the proposed method at 0--10k, the starting point of learning, is the second smallest. In this case, the method showing the lowest overlapping rate is Comp2 consisting of CommNet-based agent UAV-BSs. However, in the proposed method, the average value of the overlapping rate for all subsequent epochs is the lowest until training is complete. In addition, the overall overlapping rate of the Comp2 method is lower than the Comp1 method except for 20--30k. Therefore, communication by CommNet-based agent UAV-BSs can reduce interference as it helps agent UAV-BSs reduce overlapping rates without competing for preemption. However, although agent UAV-BSs of the Comp2 method avoid interference more efficiently than agent UAV-BSs of the Comp1 method, it converges to the second lowest QoS in Fig.~\ref{fig:total Quality} due to the lack of agent UAV-BSs' experience. Agent UAV-BSs in the Comp2 method cannot find and cover many UEs because they do not have enough discovery. Therefore, they provide mmWave communication to the smallest number of UEs compared to the proposed method and Comp2 method, as shown in Table~\ref{tab:numUEs}.

\BfPara{Energy Consumption}
Agent UAV-BSs have limited batteries, thus, this paper also investigates UAV-BSs' battery status to know whether UAV-BSs battery is completely discharged.
In the event that the UAV-BS entirely depletes its power and descends abruptly to the ground, it poses a dual risk: physical damage to the UAV-BS's hardware, and a significant safety concern due to the possibility of colliding with and injuring individuals beneath its position.
This accident implies a lack of reliability in terms of mobile access service provision.
For the evaluation, the energy quantities of untrained (initial) policy and trained policy using the proposed method are compared in the inference phase. This comparison provides validation of the training performance of the suggested approach.
The energy capacity of UAV-BS is approximately 89.224\,Wh, multiplied by the flight battery's capacity and voltage as presented in Table~\ref{tab:parameters of uav}.
Since 89.224\,W of energy per hour is available, the initial energy of the UAV-BS is 178.448\,W in an episode of 30 minutes. If the UAV-BS moves in the $x$, $y$, or $z$ direction at every step to cover moving UEs, it will fully discharge and fall to the ground before the end of the episode.
%
\begin{figure}[t!]
    \centering
    \ \ \ \ \ \includegraphics[width=0.6\linewidth]{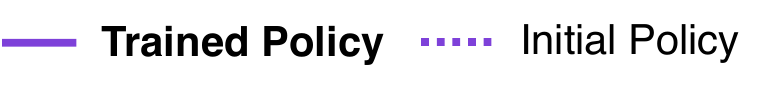}
    \subfigure[Total energy consumption.]{
    \includegraphics[width=0.45\linewidth]{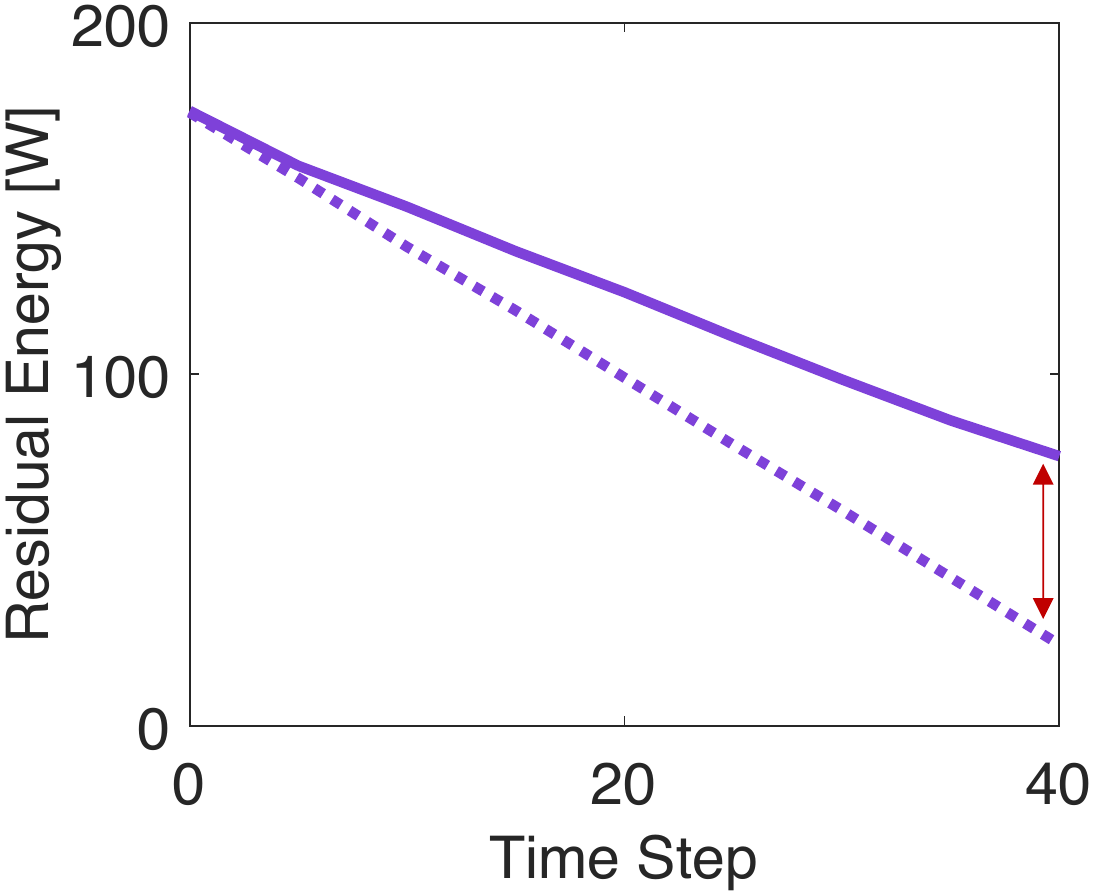}\label{fig:Energy_total}
    }
    \subfigure[Average/final residual energy.]{
    \includegraphics[width=0.45\linewidth]{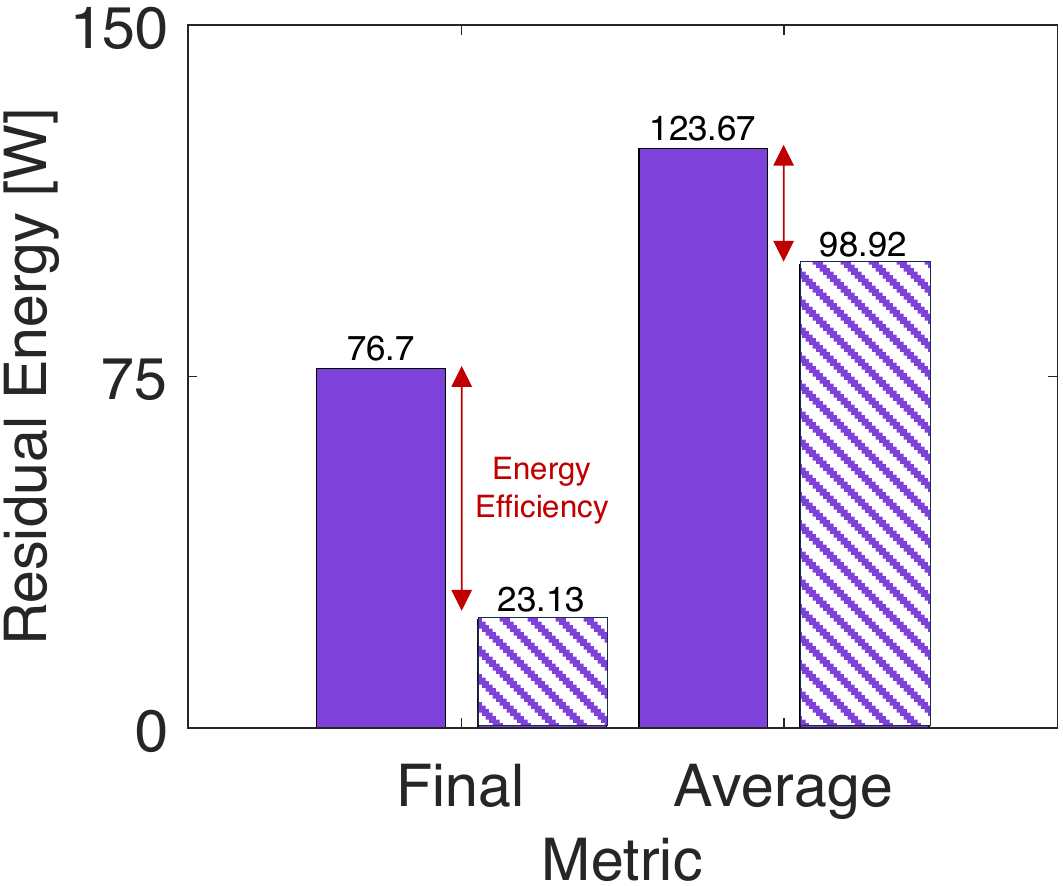}\label{fig:Energy_avg}
    }
    \caption{Residual energy of all agent UAV-BSs trained with the proposed method. Fig.~\ref{fig:Energy_total} presents the tendency of energy consumption for all agent UAV-BSs, and Fig.~\ref{fig:Energy_avg} displays both the final remaining energy of agent UAV-BSs at the end of the episode and the average residual energy throughout the progression of the episode during the inference phase.}
    \label{fig:Energy consumption}
    \vspace{-5mm}
\end{figure}
\begin{figure*}[htb]
    \centering
    \includegraphics[width=0.9\linewidth]{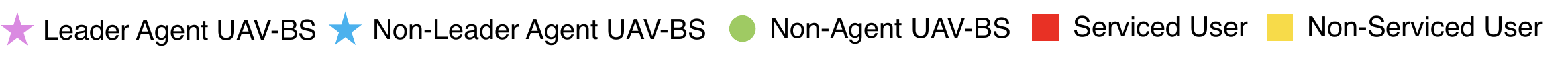}\\
    \subfigure[$t = 0$.]
    {
        \includegraphics[width=0.18\linewidth]{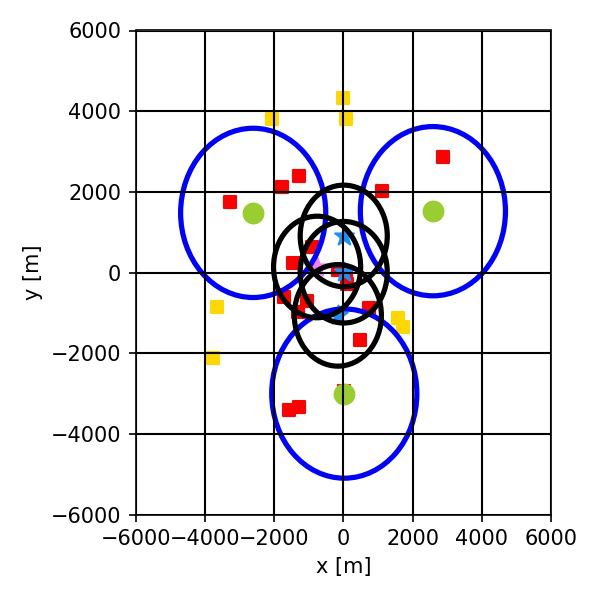}
        \label{fig:t=0}
    }
    \subfigure[$t = 5$.]
    {
        \includegraphics[width=0.18\linewidth]{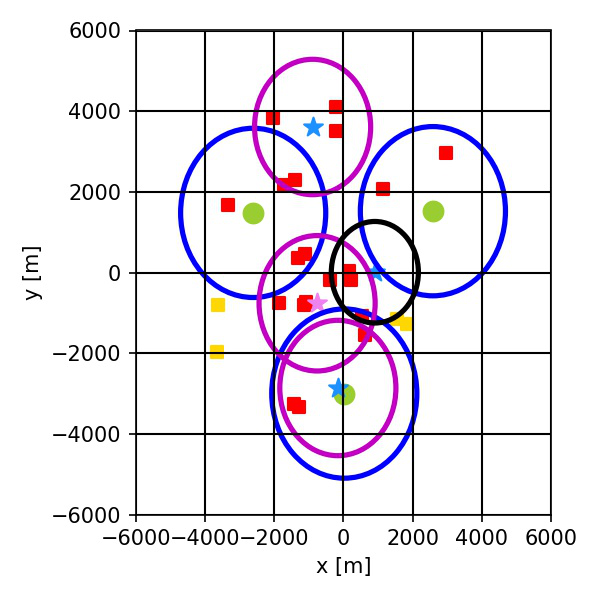}
        \label{fig:t=5}
    }
    \subfigure[$t = 10$.]
    {
        \includegraphics[width=0.18\linewidth]{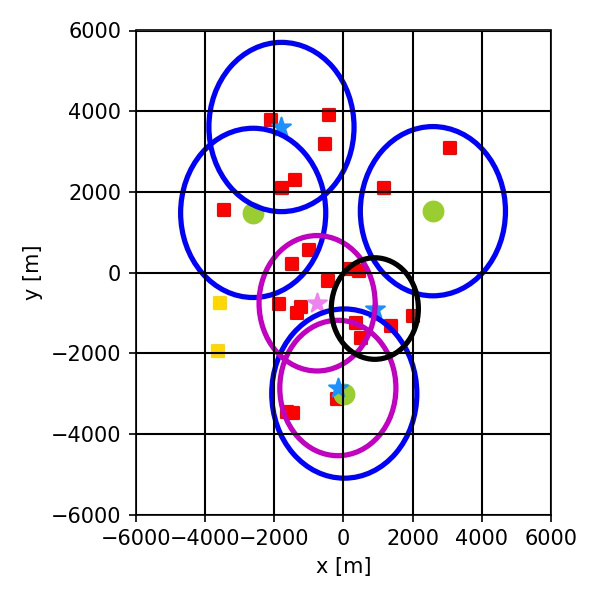}
        \label{fig:t=10}
    }
    \subfigure[$t = 15$.]
    {
        \includegraphics[width=0.18\linewidth]{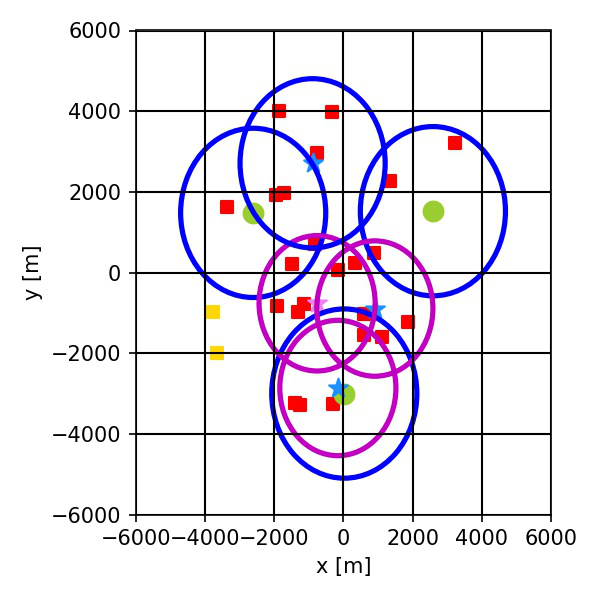}
        \label{fig:t=15}
    }
    \subfigure[$t = 20$.]
    {
        \includegraphics[width=0.18\linewidth]{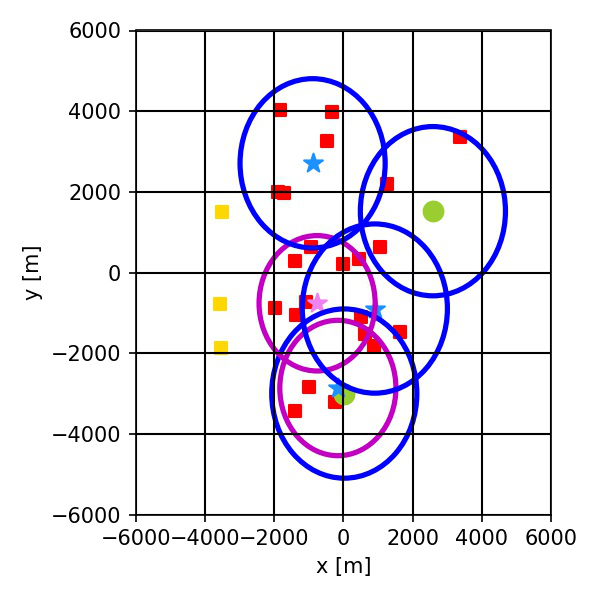}
        \label{fig:t=20}
    }\\
    \subfigure[$t = 25$.]
    {
        \includegraphics[width=0.18\linewidth]{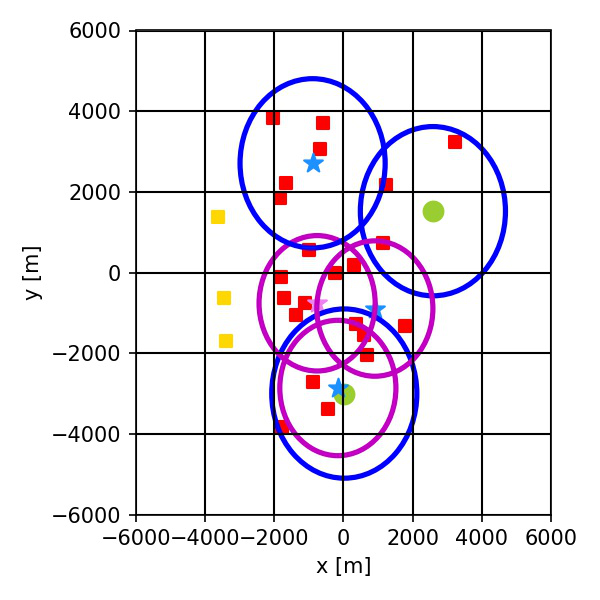}
        \label{fig:t=25}
    }
    \subfigure[$t = 30$.]
    {
        \includegraphics[width=0.18\linewidth]{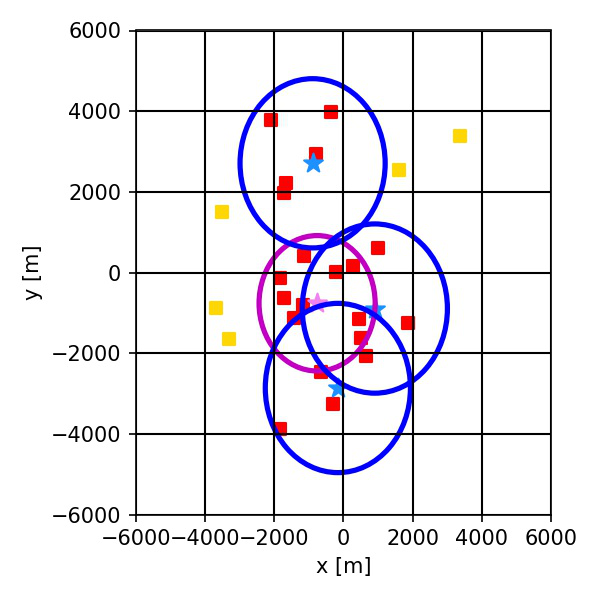}
        \label{fig:t=30}
    }
    \subfigure[$t = 35$.]
    {
        \includegraphics[width=0.18\linewidth]{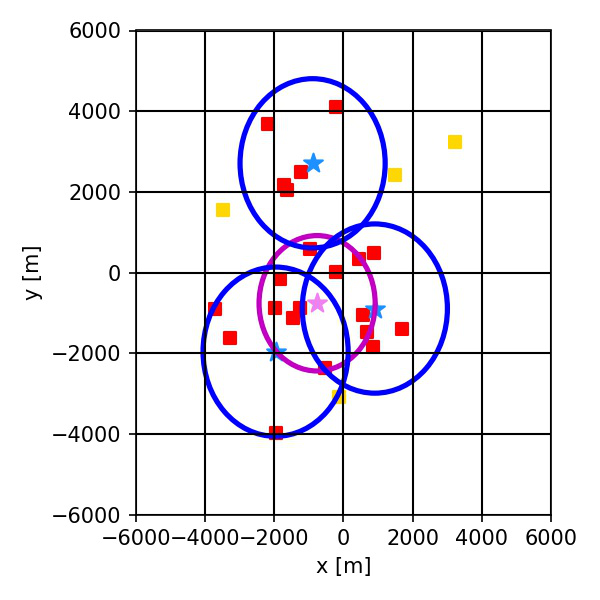}
        \label{fig:t=35}
    }
    \subfigure[$t = 40$.]
    {
        \includegraphics[width=0.18\linewidth]{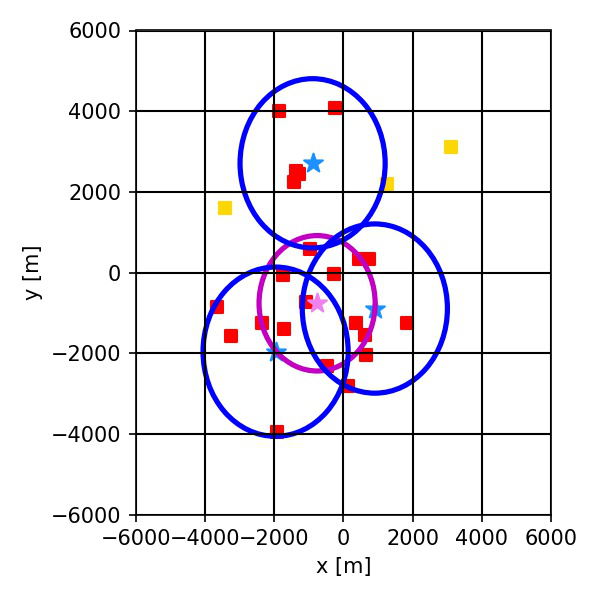}
        \label{fig:t=40}
    }
    \caption{The movement of agent UAV-BSs in the environment trained with the proposed method over time step $t$ after policy training. It is noteworthy that the adaptivity of the proposed MADRL-based positioning algorithm is verified through the movement plotting of agent UAV-BSs to the positions previously covered by malfunctioning non-agent UAV-BSs that provide wireless communication services from fixed locations.}
    \label{fig:Grid world in Proposed}
\end{figure*}
\begin{figure}
    \centering
    \ \ \ \ \ \includegraphics[width=0.75\linewidth]{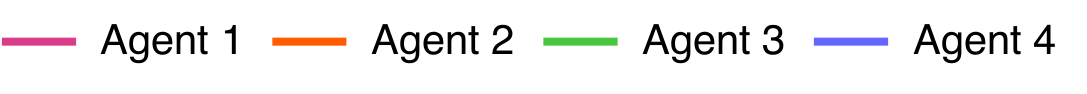}\\
    \includegraphics[width=0.9\linewidth]{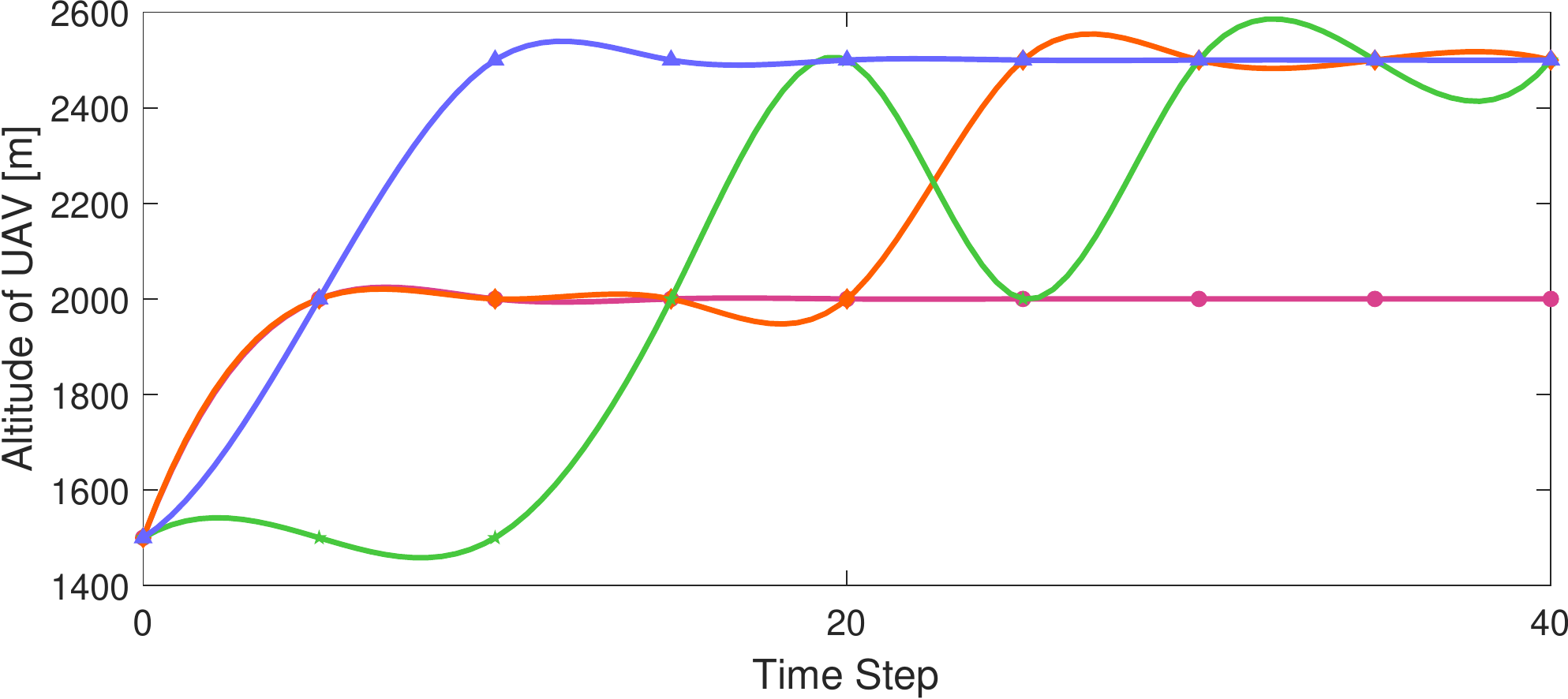}
    \caption{The altitude variation according to the UAV-BS's movement over the time step $t$ in Fig.~\ref{fig:Grid world in Proposed}.}
    \label{fig:altitude trajectory}
\end{figure}
As shown in Fig.~\ref{fig:Energy consumption}, all agent UAV-BSs initially have maximum battery capacity. The agent UAV-BSs trained with the proposed method provide mmWave services to the most significant number of UEs among all methods while avoiding discharged events.
At the end of the episode, the agent UAV-BSs operating with the trained policy have an 231.6\,\% higher residual energy than agent UAV-BSs functioning with the initial policies.
Moreover, it can be observed that agent UAV-BSs trained through the proposed method, maintain a 25.02\,\% higher average residual energy across the entire episode than agent UAV-BSs with initial policies when constructing mobile access networks.

In summary, agent UAV-BSs trained by the proposed method stably provide the highest QoS to UEs and also make the most efficient use of energy among all methods, achieved through appropriate hovering and placement. Thus, it can be inferred that UAV-BSs employing the proposed method for providing mmWave communication exhibit a high degree of service reliability.


\subsubsection{Computational Cost}

This section compares the computational cost of DNN-based and CommNet-based policies. We also calculate the computational cost of all methods using the computational cost of each policy.

\begin{table}[!t]
\centering
\caption{Comparison of computational cost with DNN-based and CommNet-based policy.}
\renewcommand{\arraystretch}{1.0}
\begin{tabular}{c||c}
\toprule[1pt]
\textbf{Metric} & \textbf{Computational Cost [FLOPS]} \\ \midrule
DNN         & 38,457  \\
CommNet     & 422,201 \\ \midrule
Proposed    & 537,572 \\ 
Comp1       & 153,828 \\
Comp2       & 1,688,804 \\
\bottomrule[1pt]
\end{tabular}
\label{tab:Computational cost}
\end{table}

Table~\ref{tab:Computational cost} shows the computational cost of the neural network based on CommNet-based and DNN-based policy in floating point operations per second (FLOPS) units~\cite{yun2022cooperative}. The CommNet-based policy requires approximately ten times higher computational cost than the DNN-based policy since the agent UAV-BS of the CommNet-based policy additionally performs the communication function with other agent UAV-BSs. Our proposed method consists of one CommNet-based agent UAV-BS that is a leader UAV-BS and three DNN-based agent UAV-BSs that are non-leader UAV-BS. Comp1 method and Comp2 method consist only of DNN-based and CommNet-based agent UAV-BSs, respectively. In a multi-agent UAV-BS environment, the more CommNet-based agent UAV-BSs than DNN-based agent UAV-BSs, the higher the computational cost in FLOPS units. This paper calculates the computational cost of each method by summing the computational costs of agent UAV-BSs present in the environment. The method we propose requires 249.5\% more and 214.2\% less computational cost, respectively, compared to the Comp1 and Comp2 methods. Note that the proposed method has the most robust user connectivity, highest service reliability, and superior reward convergence, even though the computation cost in FLOPS units is 214.2\% lower than the Comp2. Whether the proposed method requires 249.5\% more computational cost than the Comp1 method, our proposed method performs reasonably better in all respects. Therefore, our proposed method also shows unrivaled performance in computational cost.

\subsubsection{Behavior Patterns of UAV-BSs}

This section analyzes the optimal position determined by agent UAV-BSs over time after training with the proposed method. \tblue{Fig.~\ref{fig:t=0}--(i) shows the trajectory of a agent UAV-BS moving in the $x$ or $y$ direction on a 2D grid at intervals of $5$ from time $t=0$ to $t=40$, illustrating how the agent UAV-BSs cooperatively adapts in unexpected event such as the malfunction of non-agent UAV-BSs.}
Fig.~\ref{fig:altitude trajectory} shows the trajectory of a UAV-BS moving in the z-coordinate over time to provide high-quality mmWave communication to as many people as possible.
In Fig.~\ref{fig:Grid world in Proposed}, one non-agent UAV-BS malfunctions at $t\in(15, 20]$, and the others malfunction at steps $t\in(25, 30]$. Throughout the episode, agent UAV-BSs avoid competing with non-agent UAV-BSs to work together to achieve a common goal. As shown in Fig.~\ref{fig:t=0}--(d), when there is no UAV-BS malfunction, agent UAV-BSs move to areas where any UAV-BS does not provide wireless communication services to the UE. At $t\in(30, 35]$, one agent UAV-BS moves left to cover the area covered by a non-agent UAV-BS before malfunctioning at $t\in(25, 30]$. Furthermore, agent UAV-BSs remain stably in their optimal positions until the end of the episode. When the agent UAV-BS locates at a high altitude, it has a broader coverage radius according to~\eqref{eq:altitude_Agent} and~\eqref{eq:altitude_nonAgent}. However, it provides a lower quality service due to the path loss in \eqref{eq:Loss}. As shown in Fig.~\ref{fig:altitude trajectory}, as more non-agent UAV-BSs malfunction over time, more agent UAV-BSs move from lower altitudes to higher altitudes over time. These behavior patterns of agent UAV-BSs are reasonable because they can serve as many UEs as possible. \tblue{Through the experiment, our proposed method proves the ability of agent UAV-BSs that effectively and adaptively operates with the malfunction event of existing non-agent UAV-BSs.}

\section{Concluding Remarks}\label{sec:con}
This paper proposes a novel cooperative positioning algorithm for multiple UAV-BSs in the context of ITS. The proposed algorithm lies in the concepts of MADRL/CommNet. The primary objective is to achieve reliable V2X communication services in ITS, while ensuring dependable operations.
For the proposed neural network training, optimization criteria is defined based on the reward functions of MADRL/CommNet formulation, \textit{i.e.}, service reliability with QoS requirements and energy-efficiency. Moreover, 60\,GHz mmWave wireless access is also utilized in order to realize the benefits of \textit{i)} ultra-wide-bandwidth for high-speed communications and \textit{ii)} high-directional communications for spatial reuse.
Lastly, the performance of our proposed MADRL/CommNet-based multi-UAV-BS positioning algorithm is evaluated in various ways with data-intensive simulations; and the superiority of the proposed algorithm over other existing algorithms has been confirmed in terms of service quality and reliability.

\bibliographystyle{IEEEtran}
\bibliography{ref_mobilebs,ref_aimlab}

\begin{IEEEbiography}[{\includegraphics[width=1in,height=1.25in,clip,keepaspectratio]{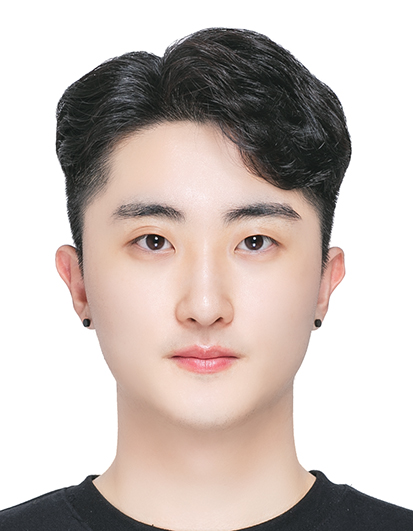}}]{Chanyoung Park} 
is currently a Ph.D. student at the Department of Electrical and Computer Engineering, Korea University, Seoul, Republic of Korea, since September 2022. He received the B.S. degree in electrical and computer engineering from Ajou University, Suwon, Republic of Korea, in 2022, with honor (early graduation). His research focuses include deep learning algorithms and their applications to communications and networks. 
\end{IEEEbiography}

\begin{IEEEbiography}[{\includegraphics[width=1in,height=1.25in,clip,keepaspectratio]{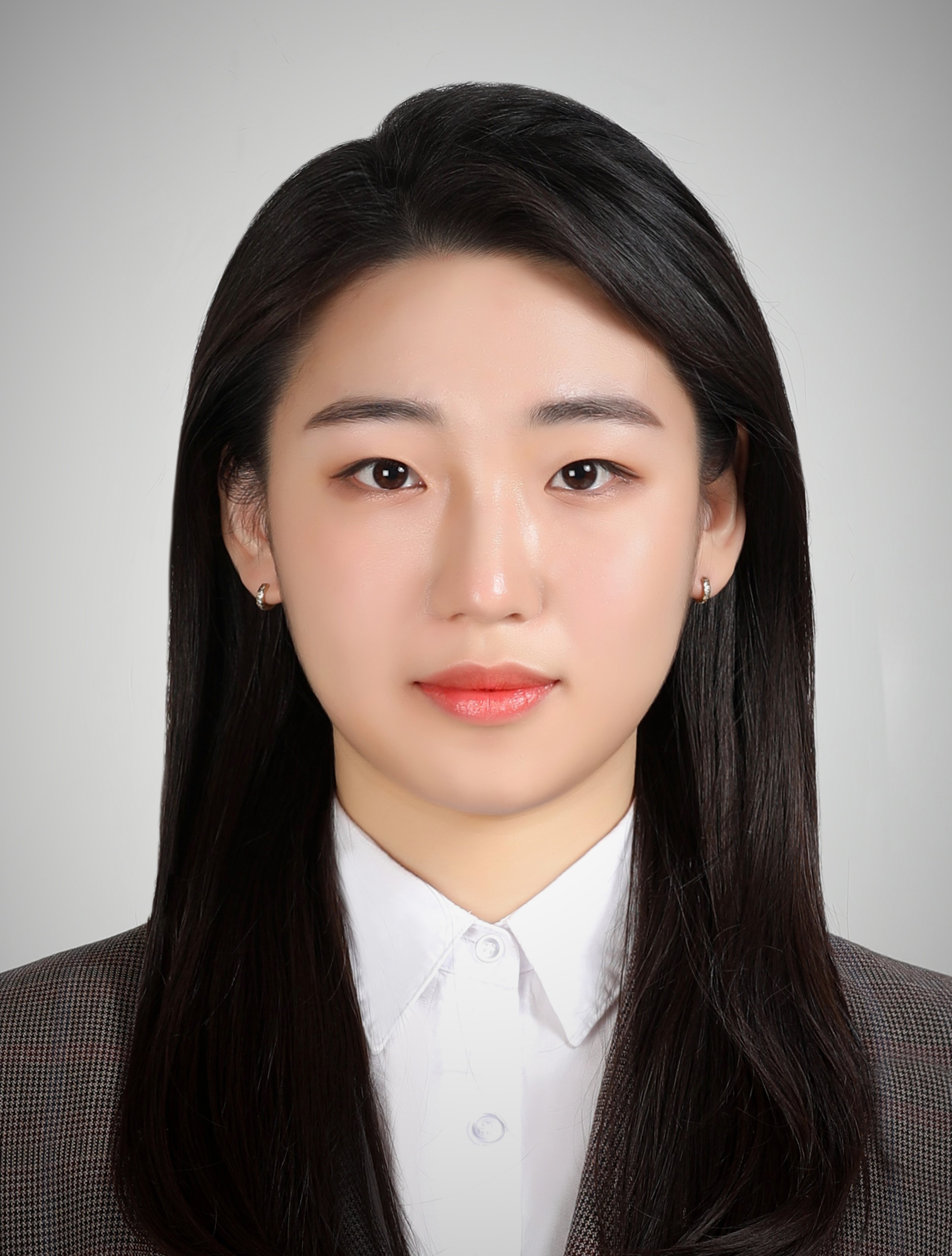}}]{Soohyun Park} has been a postdoctoral scholar at the Department of Electrical and Computer Engineering, Korea University, Seoul, Republic of Korea, since 2023. 
She received her Ph.D. degree in electrical and computer engineering at Korea University, Seoul, Republic of Korea, in 2023. She also received her B.S. degree in computer science and engineering from Chung-Ang University, Seoul, Republic of Korea, in 2019. Her research focuses include deep learning algorithms and their applications to computer networking, autonomous mobility platforms, and quantum multi-agent distributed autonomous systems. 

She was a recipient of the IEEE Vehicular Technology Society (VTS) Seoul Chapter Award (2019), Bronze Paper Award by IEEE Seoul Section Student Paper Contest (2020), and Best Reviewer Award by \textit{ICT Express (Elsevier)} (2021).
\end{IEEEbiography}

\begin{IEEEbiography}[{\includegraphics[width=1in,height=1.25in,clip,keepaspectratio]{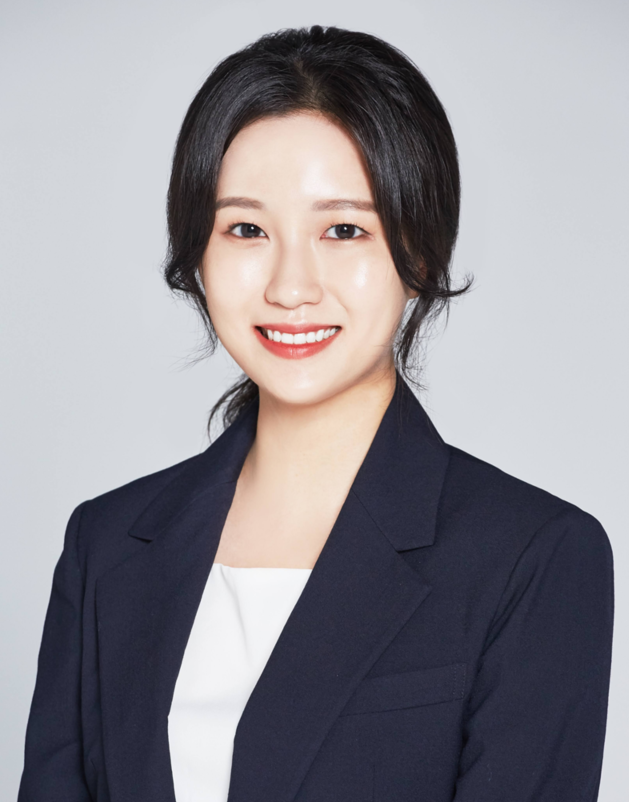}}]{Soyi Jung} (Member, IEEE) has been an assistant professor at the Department of Electrical of Computer Engineering, Ajou University, Suwon, Republic of Korea, since September 2022. Before joining Ajou University, she was an assistant professor at Hallym University, Chuncheon, Republic of  Korea, from 2021 to 2022; a visiting scholar at Donald Bren School of Information and Computer Sciences, University of California, Irvine, CA, USA, from 2021 to 2022; a research professor at Korea University, Seoul, Republic of Korea, in 2021; and a researcher at Korea Testing and Research (KTR) Institute, Gwacheon, Republic of Korea, from 2015 to 2016. She received her B.S., M.S., and Ph.D. degrees in electrical and computer engineering from Ajou University, Suwon, Republic of Korea, in 2013, 2015, and 2021, respectively. 
She was a recipient of Best Paper Award by KICS (2015), Young Women Researcher Award by WISET and KICS (2015), Bronze Paper Award from IEEE Seoul Section Student Paper Contest (2018), ICT Paper Contest Award by Electronic Times (2019), and IEEE ICOIN Best Paper Award (2021).
\end{IEEEbiography}

\begin{IEEEbiography}[{\includegraphics[width=1in,height=1.25in,clip,keepaspectratio]{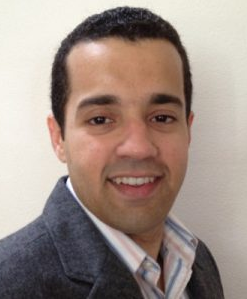}}]{Carlos Cordeiro} (SM’09--F'19)
is a wireless CTO at Intel Corporation, where he is responsible for next generation wireless connectivity technology strategy, standards, ecosystem engagements, and regulatory. In the Wi-Fi Alliance, he is a member of the Board of Directors and serves as its Technical Advisor. 

Due to his contributions to wireless communications, Dr. Cordeiro received several awards including the prestigious Intel Inventor of the Year Award in 2016, the 2017 IEEE Standards Medallion, the IEEE Outstanding Engineer Award in 2011, and the IEEE New Face of Engineering Award in 2007. 

Dr. Cordeiro is the co-author of two textbooks on wireless published in 2006 and 2011, has published over 110 papers in the wireless area alone, and holds over 250 patents. He is the associate editor-in-chief of the \textsc{IEEE Communications Standards Magazine}, and has served as Editor of various journals including the \textsc{IEEE Transactions on Mobile Computing}, the \textsc{IEEE Journal on Selected Areas in Communications}, the \textsc{IEEE Transactions on Wireless Communications}, the \textsc{IEEE Wireless Communication Letters} and the \textsc{ACM Mobile Computing and Communications Review}. He has also served in the leadership and technical program committee of numerous conferences. 

Dr. Cordeiro is an IEEE Fellow.
\end{IEEEbiography}

\begin{IEEEbiography}[{\includegraphics[width=1in,height=1.25in,clip,keepaspectratio]{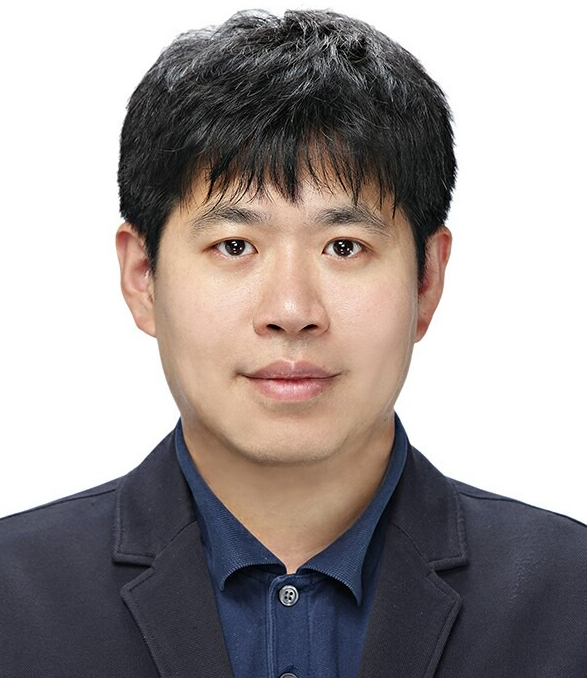}}]{Joongheon Kim}
(Senior Member, IEEE) has been with Korea University, Seoul, Korea, since 2019, where he is currently an associate professor at the Department of Electrical and Computer Engineering and also an adjunct professor at the Department of Communications Engineering (co-operated by Samsung Electronics) and the Department of Semiconductor Engineering (co-operated by SK Hynix). He received the B.S. and M.S. degrees in computer science and engineering from Korea University, Seoul, Korea, in 2004 and 2006; and the Ph.D. degree in computer science from the University of Southern California (USC), Los Angeles, CA, USA, in 2014. Before joining Korea University, he was a research engineer with LG Electronics (Seoul, Korea, 2006--2009), a systems engineer with Intel Corporation Headquarter (Santa Clara in Silicon Valley, CA, USA, 2013--2016), and an assistant professor of computer science and engineering with Chung-Ang University (Seoul, Korea, 2016--2019). 

He serves as an editor and guest editor for \textsc{IEEE Transactions on Vehicular Technology}, \textsc{IEEE Transactions on Machine Learning in Communications and Networking}, \textsc{IEEE Communications Standards Magazine}, \textit{Computer Networks (Elsevier)}, and \textit{ICT Express (Elsevier)}. He is also a distinguished lecturer for \textit{IEEE Communications Society (ComSoc)} and \textit{IEEE Systems Council}. He is an executive director of the Korea Institute of Communication and Information Sciences (KICS). 

He was a recipient of Annenberg Graduate Fellowship with his Ph.D. admission from USC (2009), Intel Corporation Next Generation and Standards (NGS) Division Recognition Award (2015), 
\textsc{IEEE Systems Journal} Best Paper Award (2020), IEEE ComSoc Multimedia Communications Technical Committee (MMTC) Outstanding Young Researcher Award (2020), IEEE ComSoc MMTC Best Journal Paper Award (2021), Best Special Issue Guest Editor Award by \textit{ICT Express} (2022), and Best Editor Award by \textit{ICT Express} (2023). He also received several awards from IEEE conferences including IEEE ICOIN Best Paper Award (2021), IEEE Vehicular Technology Society (VTS) Seoul Chapter Awards (2019, 2021, and 2022), and IEEE ICTC Best Paper Award (2022). 
\end{IEEEbiography}
\end{document}